\documentclass{article}

\PassOptionsToPackage{numbers, compress}{natbib}

\usepackage[preprint]{neurips_2025}

\usepackage[utf8]{inputenc} 
\usepackage[T1]{fontenc}    
\usepackage{hyperref}       
\usepackage{url}            
\usepackage{booktabs}       
\usepackage{amsfonts}       
\usepackage{nicefrac}       
\usepackage{microtype}      
\usepackage{xcolor}         

\usepackage{subfigure}
\usepackage{graphicx}
\usepackage{enumitem}

\usepackage{amsmath}
\usepackage{amssymb}
\usepackage{mathtools}
\usepackage{amsthm}
\usepackage{multirow}

\usepackage{algorithm}
\usepackage{algorithmic}

\usepackage{booktabs} 

\title{Long-Tailed Out-of-Distribution Detection with Refined Separate Class Learning}

\author{%
  Shuai Feng  \\
 Nanjing Agricultural University\\
  shuaifeng@njau.edu.cn\\
  \And
  Yuxin Ge \\
  Nanjing University \\
  yuxinge@smail.nju.edu.cn\\
  \AND
  Yuntao Du \\
  Shandong University \\
  yuntaodu@sdu.edu.cn\\
  \And
  Mingcai Chen \\
  Nanjing University of Posts and Telecommunications\\
  chenmc@njupt.edu.cn\\
  \And
  Chongjun Wang \\
  Nanjing University \\
  chjwang@nju.edu.cn\\
  \And
  Lei Feng \thanks{Corresponding author}\\
  Southeast University \\
  fenglei@seu.edu.cn\\
}

\begin{document}

\maketitle

\begin{abstract}
 Out-of-distribution (OOD) detection is crucial for deploying robust machine learning models. However, when training data follows a long-tailed distribution, the model’s ability to accurately detect OOD samples is significantly compromised, due to the confusion between OOD samples and head/tail classes. 
To distinguish OOD samples from both head and tail classes, the separate class learning (SCL) approach has emerged as a promising solution, which separately conducts head-specific and tail-specific class learning.
To this end, we examine the limitations of existing works of SCL and
reveal that the OOD detection performance is sensitive to static temperature value and subject to the impact of informative outliers.
To mitigate these limitations, we propose a novel approach termed \textbf{R}efined \textbf{S}eparate \textbf{C}lass \textbf{L}earning (\textbf{RSCL}), which leverages dynamic class-wise temperature adjustment to modulate the temperature parameter for each in-distribution class and utilizes informative outlier mining to identify diverse types of outliers based on their affinity with head and tail classes. Extensive experiments demonstrate that RSCL achieves superior OOD detection performance while improving the classification accuracy on in-distribution data. 
\end{abstract}

\section{Introduction}
\label{sec:intro}

Deep neural networks (DNNs) are widely known for generating overconfidence predictions when encountering unfamiliar data, leading to the misclassification of out-of-distribution (OOD) samples from unknown classes as members of known classes with high confidence \cite{nguyen2015deep,yang2024generalized}. Such behavior can yield detrimental consequences in safety-critical applications such as autonomous driving, medical diagnosis, and fraud detection \cite{leibig2017leveraging,kendall2017uncertainties,fanai2023novel}. Hence, the task of OOD detection, which focuses on accurately identifying OOD test data while maintaining high performance on in-distribution (ID) data, has received significant attention. 

While existing state-of-the-art OOD detection methods \cite{MSP,odin,maha,energy,case} have demonstrated remarkable success when evaluated on class-balanced ID training data, they suffer from significant performance deterioration when the training sets follow a long-tailed distribution \cite{pascl}. In such distributions, a few majority categories (head classes) dominate the dataset, while most minority categories (tail classes) have limited samples available. The primary challenges in long-tailed OOD detection arise from the confusion between OOD samples and head/tail classes, which manifests in two ways: (1) tail-class samples are erroneously treated as OOD samples \cite{pascl,TSCL,EAT,COCL}, and (2) OOD samples are often incorrectly classified into head classes \cite{COCL}. Hence, it is imperative to effectively distinguish OOD samples from both head and tail classes. 
The separate class learning (SCL) approach has emerged as a promising solution, designed to scrutinize tail and head samples separately through tail-specific and head-specific class learning mechanisms.
However, our analysis reveals two persistent limitations in SCL:

\begin{figure}[] 
	\centering  
	\subfigure[\small Influence of static temperature]{
		\label{fig1a}
	\includegraphics[width=0.49\linewidth]{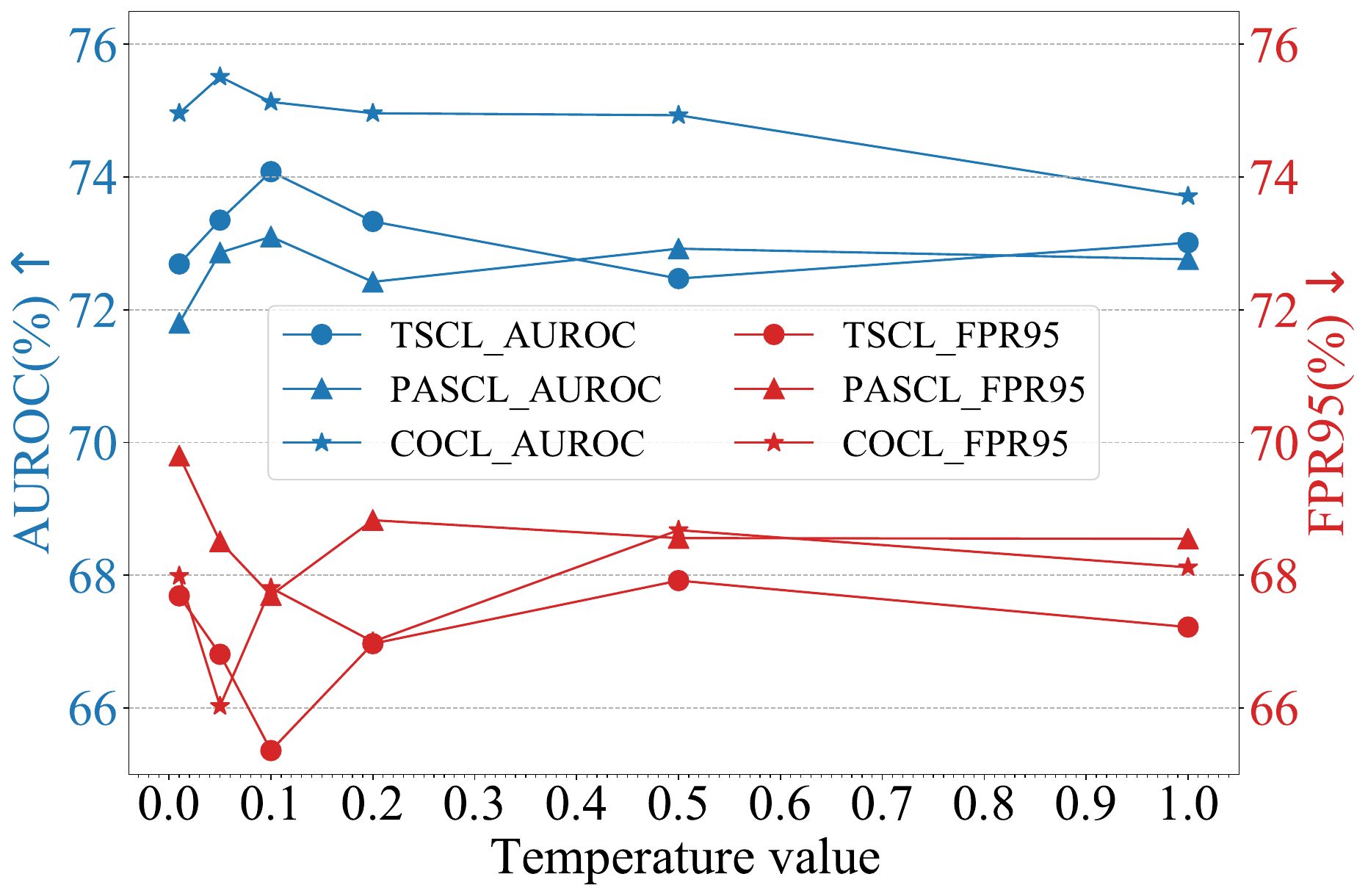}}
       \subfigure[\small Advantage of informative outlier mining]{
		\label{fig1b}
	\includegraphics[width=0.455\linewidth]{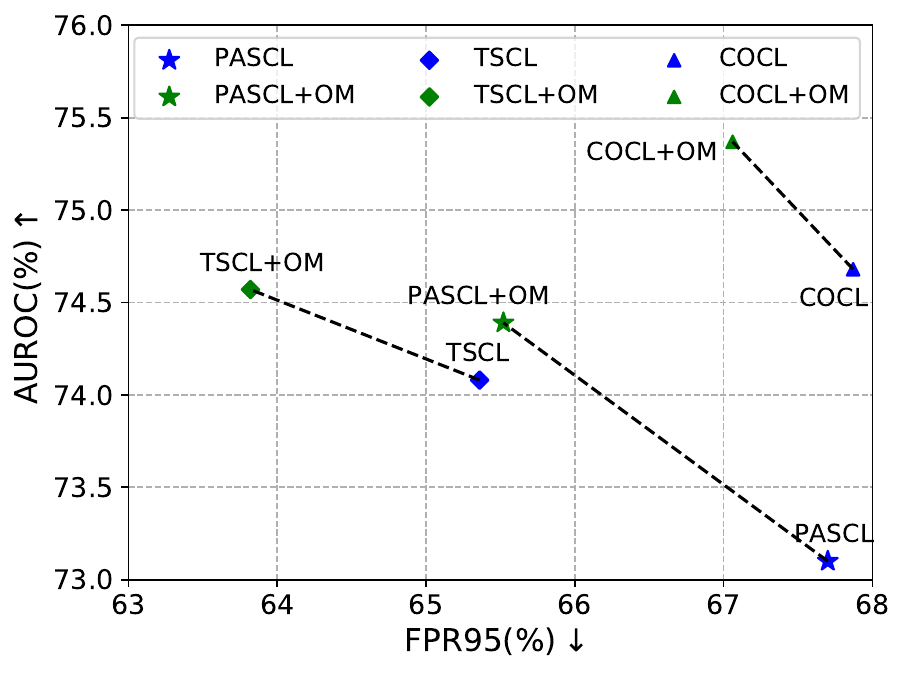}}
    \caption{The OOD detection performance of representative methods (PASCL \cite{pascl}, TSCL \cite{TSCL}) and COCL \cite{COCL} can be influenced by (a) static temperature values, and (b) informative outlier mining (OM) from the auxiliary OOD training data. The experiments are conducted on CIFAR100-LT benchmark. The AUROC and FPR95 are averaged over six OOD test sets (refer to Section \ref{Experiment Settings}).}
	\label{figure1}
    \vspace{-5mm}
\end{figure}

1) \textbf{Sensitive to static temperature}: To distinguish OOD samples from head/tail classes, prevalent works predominantly rely on diverse augmentation within the realm of  supervised contrastive learning \cite{pascl,TSCL} or utilize learnable prototypes for tail classes \cite{TSCL,COCL} to push head/tail samples away from OOD samples. Typically, the temperature parameter, which plays a pivotal role in regulating how much attention the model pays to samples, is employed as a static value across anchor, positive, and negative samples for simplicity.
However, as illustrated in Figure \ref{fig1a}, the OOD detection performance of these methods is influenced by varying temperature values.
This sensitivity can be attributed to the use of a static temperature value, which results in uniform pulling and uniform pushing, a configuration that is incongruent with the intricacies of long-tailed data distribution. It is more reasonable to dynamically adjust the temperature value of each head and tail class to better align with the imbalanced distribution of class samples inherent in the training data.

2) \textbf{Without consideration of inforamtive outliers}: Recent works \cite{pascl, EAT, COCL, TSCL} focus on improving long-tailed OOD detection performance through an approach called outlier exposure (OE) \cite{OE} that utilizes an unlabeled auxiliary training set as OOD training data.  
However, these methods have primarily relied on a random selection of outliers from the auxiliary OOD training data. 
This indiscriminate selection approach may lead to the inclusion of numerous uninformative outliers that lack significant alignment with either the head or tail classes. Such outliers exhibit restricted OOD characteristics, potentially yielding marginal enhancements to the model's performance \cite{atom}. As illustrated in Figure \ref{fig1b}, utilizing informative outliers mined from the OOD training data (refer to Section \ref{Informative Outlier Mining}) can significantly improve the OOD detection performance. This highlights the advantages of enhancing the quality of auxiliary outliers.

To comprehensively address the aforementioned limitations of SCL, we introduce \textbf{A}daptable \textbf{T}ail class \textbf{S}upervised \textbf{C}ontrastive \textbf{L}earning (A-TSCL) and \textbf{O}OD-prototype-aware \textbf{H}ead class \textbf{L}earning (A-OHL) as integral components of tail-specific and head-specific class learning, respectively. To mitigate these identified limitations, we propose \textbf{R}efined \textbf{S}eparate \textbf{C}lass learning (RSCL). Specifically, departing from the conventional static temperature value application across all ID classes, we propose dynamic class-wise temperature adjustment to modulate the temperature parameter for each head and tail class. This adaptive approach enables varying degrees of pulling within each tail class and distinct levels of pushing between OOD samples and head classes. 
In addition, as opposed to random outlier selection from the auxiliary OOD training set, we propose informative outlier mining to identify diverse types of outliers based on their affinity with head and tail classes. This will help to further enhance the efficacy of separate class learning. Through the above refinements to separate class learning, OOD samples are less prone to confusion with head and tail classes, thus improving the long-tailed OOD detection performance. 

In conclusion, our main contributions are as follows:

\begin{itemize}
 \item We analyze the underlying limitations of the separate class learning approach for long-tailed OOD detection and reveal that the OOD detection performance is notably influenced by the presence of uninformative outliers and the use of static temperature parameters.
\item We propose refined separate class learning (RSCL), a novel approach for long-tailed OOD detection. RSCL utilizes dynamic class-wise temperature adjustment and informative outlier mining to mitigate the above limitations and effectively distinguish OOD samples from both head and tail classes.
\item Extensive evaluation and ablation studies demonstrate that RSCL achieves superior OOD detection performance while improving ID classification accuracy.
\end{itemize}

\section{Related Work} 
\label{sec2}
\textbf{OOD Detection.} \quad 
Numerous methods have been developed to tackle OOD detection challenges within the realm of computer vision. One prevalent approach involves the utilization of post-hoc inference methods \cite{MSP,maha,energy,odin,GradNorm,KNN,vim,ash,SHE,rankfeat,Submanifold,yuandiscriminability}. These methods primarily focus on devising score functions that depend on features, logits, or softmax probabilities. Another approach involves training methods that do not rely on outlier data \cite{godin,logitnorm,cider,case,csi}. These methods incorporate training-time regularization techniques to enhance the OOD detection capabilities. The third approach entails training methods that utilize outlier data, assuming access to auxiliary OOD training samples \cite{OE,energy,OECC,SOFL,DAC,atom,vos,npos,nie2024out,wangout,zheng2023out,lu2023uncertainty}. However, it is crucial to note that all these methods have primarily been developed and evaluated on datasets with balanced ID training data, thereby potentially exhibiting limitations when applied to imbalanced ID datasets.

\textbf{Long-tailed OOD Detection.}
\quad Compared with OOD detection on balanced ID training data, significantly less work has been done on long-tailed scenarios. The pioneering work, PASCL \cite{pascl}, formulates the problem  and reveals that simple combinations of existing OOD detection and long-tailed recognition methods \cite{re_weighting,decoupling,LA} achieve unsatisfactory performance. PASCL identifies the challenge arising from the models' struggle to differentiate between tail-class and OOD samples. 
To tackle this challenge, EAT \cite{EAT} introduces dynamic virtual labels for OOD samples and augments the context-limited tail classes. TSCL \cite{TSCL} introduces an extra rejection class label for auxiliary OOD training data. 
COCL \cite{COCL} recently demonstrates that OOD samples are often misclassified into head classes and proposes a novel approach called calibrated outlier class learning to distinguish OOD samples from both head and tail classes. 
However, these methods commonly adopt a random selection of outliers from the auxiliary OOD training data.  We instead utilize informative outlier mining to identify and utilize diverse types of outliers based on their affinity with head and tail classes. 

\textbf{Supervised Contrastive Learning.} \quad Supervised contrastive learning \cite{supcon} has recently shown strong potential in enhancing long-tailed OOD detection. In its original formulation, all classes play the same role that pulling is done with every class and pushing between every pair of different classes. To distinguish OOD samples from tail classes, PASCL \cite{pascl} proposes partial and asymmetric supervised contrastive learning, restricting contrastive interactions to  tail classes and OOD samples. Building on this idea,  TSCL \cite{TSCL} introduces tail-class prototype induced supervised contrastive learning to improve the performance of long-tail OOD detection further. Both PASCL and TSCL impose uniform pulling within each tail class and uniform pushing between OOD samples and tail classes, due to the employment of a static temperature value.
In contrast, our approach employs dynamic class-wise temperature adjustment to modulate the temperature parameter for both head and tail class. 

\begin{figure}[ht]
  \centering
   \includegraphics[width=1.0\linewidth]{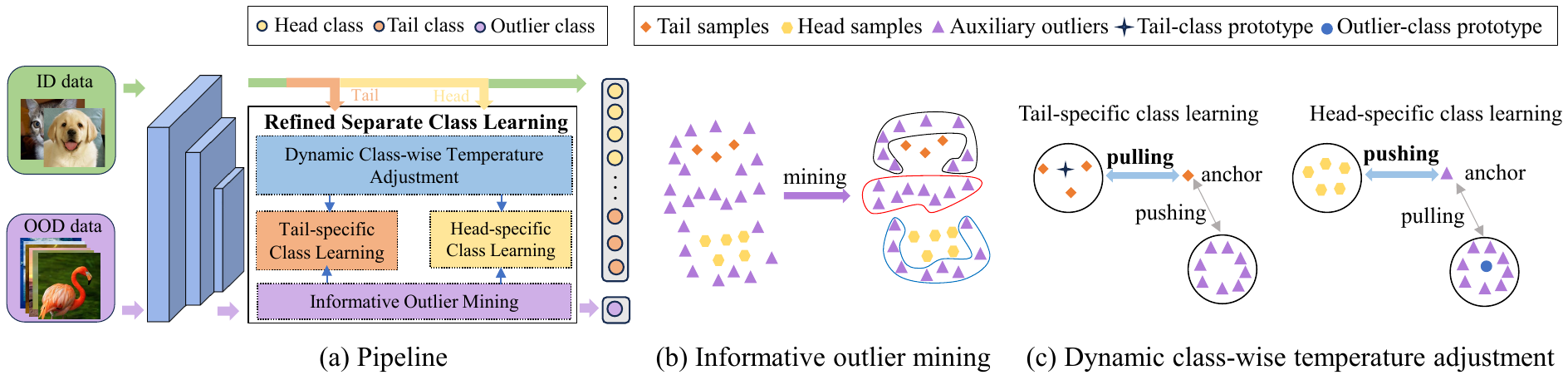}
   \vspace{-3mm}
   \caption{Overview of our approach RSCL. (a) High-level architecture of RSCL components. (b) Conceptual diagram of informative outlier mining: outliers are categorized as $\emph{Tail-class-like OOD}$ (black boundary), $\emph{Neutral OOD}$ (red boundary), and $\emph{Head-class-like OOD}$ (blue boundary). (c) Conceptual illustration of dynamic temperature adjustment for tail-specific class learning through A-TSCL (left diagram) and head-specific class learning via A-OHL (right diagram).}

   
   \label{fig2}
\end{figure}
\section{Preliminaries}
\textbf{Problem Statement.} \quad
Let $\mathcal{X}={X^{\text {in}} \cup X^{\text {out}} }$ represents the input space and $\mathcal{Y^{\text {in}}}=\left\{1,2,...,C\right\}$ be the set of $C$ imbalanced ID classes in the label space. The objective of long-tailed OOD detection is to learn a classifier $f$ that for any test data $\boldsymbol{x} \in \mathcal{X}$: if $\boldsymbol{x}$ is drawn from $X^{\text {in}}$ (from either head or tail classes), $f$ can classify $\boldsymbol{x}$ into the correct ID class; and if $\boldsymbol{x}$ is drawn from $X^{\text {out}}$, $f$ can identify $\boldsymbol{x}$ as OOD data. Typically, genuine OOD data is unavailable during the training phase, as OOD samples are unknown instances. However, auxiliary samples that do not belong to $X^{\text {out}}$ but are drawn from a different distribution other than $X^{\text {in}}$ are often accessible. These auxiliary samples can be utilized as pseudo-OOD samples to improve OOD detection performance and ID classification accuracy.

\textbf{Outlier Class for Auxiliary OOD Training Data.} \quad
Outlier Exposure (OE) approach \cite{OE} is widely adopted for OOD detection on balanced ID datasets by utilizing an unlabeled auxiliary training set as OOD training data. However, when the training set follows a long-tailed distribution, the commonly adopted uniform distribution in OE becomes an ineffective prior. To address this issue, a straightforward solution is to learn an outlier class that encompasses the OOD samples \cite{COCL,TSCL}. Specifically, for a classification problem with $C$ classes, the label space is extended by explicitly adding a separate class $(C + 1)$ as an outlier class. During training, both the ID data $\mathcal{D}^{\text{train}}_{\text{in}}=(X_{\text{in}}, Y_{\text{in}})$  and auxiliary data $\mathcal{D}^{\text{train}}_{\text{out}}=(X_{\text{out}}, C+1)$  are utilized. The training objective is defined as follows:
\begin{equation}
\label{eq1}
\mathcal{L}_{\text{OCL}}=\mathbb{E}_{(\boldsymbol{x}, y) \sim \mathcal{D}^{\mathrm{train}}_{\mathrm{in}}}[\ell(f(\boldsymbol{x}), y]+\alpha \mathbb{E}_{\boldsymbol{x} \sim \mathcal{D}^{\mathrm{train}}_{\mathrm{out}}}[\ell(f(\boldsymbol{x}), \tilde{y}],
\end{equation}
where  $f(\boldsymbol{x})$ is the logits output, $\tilde{y} = C+1$, $\ell$ is a cross entropy loss, and $\alpha$ is a hyper-parameter.

\section{Refined Separate Class Learning}

We introduce a novel  approach, termed \textbf{R}efined \textbf{S}eparate \textbf{C}lass \textbf{L}earning (\textbf{RSCL}), designed to mitigate the limitations in SCL. As shown in Figure \ref{fig2}(a), RSCL utilizes dynamic class-wise temperature adjustment and informative outlier mining to facilitate tail-specific and head-specific class learning.  The components in RSCL are jointly trained with a shared feature extractor, ensuring comprehensive learning of representative features for all classes. Below we introduce each component in detail.

\subsection{Separate Class Learning}
\label{Separate Class Learning}
\textbf{Tail-specific Class Learning.}
\quad Due to the scarcity of tail-class samples in training data, models often mistakenly treat them as OOD samples \cite{pascl,EAT,COCL,TSCL}. To mitigate this issue, an \textbf{A}daptable \textbf{T}ail class \textbf{S}upervised \textbf{C}ontrastive \textbf{L}earning (A-TSCL) is utilized to effectively distinguish tail classes from OOD samples. The formulation is as follows:
\begin{equation}
\label{eq2}
\begin{aligned}
& \mathcal{L}_{\text {tail }}=\mathbb{E}_{\boldsymbol{x} \sim \mathcal{I}}\left[\mathcal{L}_{\text{A-TSCL}}(\boldsymbol{x})\right], \text { where } \mathcal{L}_{\text{A-TSCL}}(\boldsymbol{x})= \\
& \sum\limits_{\boldsymbol{p} \in \mathcal{P}(\boldsymbol{x})}-\frac{1}{|\mathcal{P}(\boldsymbol{x})|} \log \frac{\exp (\frac{z(\boldsymbol{x})^T z(\boldsymbol{p})} {\hat{\tau}_{c(\boldsymbol{x})}})}{\sum\limits_{\boldsymbol{t}\in \mathcal{B}_\text{t} \backslash\{\boldsymbol{x}\}} \exp (\frac{z(\boldsymbol{x})^T z(\boldsymbol{t})}  {\hat{\tau}_{c(\boldsymbol{x})}}) +{\sum\limits_{\tilde{\boldsymbol{x}} \in \tilde{\mathcal{B}}} \exp (\frac{z(\boldsymbol{x})^T z(\tilde{\boldsymbol{x}})} {\tau})}}.
\end{aligned}
\end{equation}
where $\mathcal{P}(\boldsymbol{x})=\left\{\boldsymbol{p} \mid y(\boldsymbol{p})=y(\boldsymbol{x}), \boldsymbol{p} \in \mathcal{B} \backslash\{\boldsymbol{x}\}\right\}$ denotes a set of positive samples sharing the same ground-truth label as the tail sample $\boldsymbol{x}$, 
$\mathcal{B}$ is a batch of ID training samples from $\mathcal{D}^{\text{train}}_{\text{in}}$, 
$\mathcal{B}_\text{t}$ is a set of tail-class samples within $\mathcal{B}$,
$\tilde{\mathcal{B}}$ is a batch of informative outliers from $\mathcal{D}^{\text{train}}_{\text{out}}$, 
 $z(\boldsymbol{x})$ represents the output of a non-linear projection head on the model’s penultimate layer for sample $\boldsymbol{x}$, $\hat{\tau}_{c(\boldsymbol{x})}$ indicates a dynamic temperature parameter tailored for a tail class that is explicitly defined in Eq.(\ref{eq4}), and $\mathcal{I} = \mathcal{B}_\text{t}\cup \mathcal{W}_{\text {t}}$ with $\mathcal{W}_t = (\boldsymbol{Z}_{t+1},\boldsymbol{Z}_{t+2},...,\boldsymbol{Z}_C) \in \mathbb{R}^{(C-t) \times D}$ being a set of tail-class prototypes. Following TSCL \cite{TSCL}, $\mathcal{W}_{\text {t}}$ is defined as trainable tail-class weight parameters of the model. 
 To obtain $\mathcal{W}_t$, we first extract the tail-class weights $\mathcal{T}_t = (\boldsymbol{W}_{t+1},\boldsymbol{W}_{t+2},...,\boldsymbol{W}_C) \in \mathbb{R}^{(C-t) \times D}$ from the classifier weights $\mathcal{W} = (\boldsymbol{W}_1,\boldsymbol{W}_2,...,\boldsymbol{W}_C) \in \mathbb{R}^{C \times D}$. Subsequently, the tail-class weights in $\mathcal{T}_t$ are separately transformed by a multilayer perceptron (MLP). Finally, $\mathcal{W}_t$ are obtained after applying $\ell_2$ normalization.

The main difference between A-TSCL and previous works \cite{pascl,EAT,COCL,TSCL} is the dynamic temperature parameter $\hat{\tau}_{c(\boldsymbol{x})}$, which is designed to adjust the temperature between tail samples during training. 

\textbf{Head-specific Class Learning.} 
\quad Given the prevalent abundance of head-class samples in the training data, models often exhibit a prominent bias towards head classes when detecting OOD samples \cite{COCL}. Consequently, OOD samples are frequently misclassified into head classes. \textbf{A}daptable \textbf{O}OD-prototype-aware \textbf{H}ead class \textbf{L}earning (A-OHL) is defined to effectively differentiate head classes from OOD samples. The formulation is as follows:
\begin{equation}
\label{eq3}
\begin{aligned}
& \mathcal{L}_{\text {head}}=\mathbb{E}_{\tilde{\boldsymbol{x}} \sim \tilde{\mathcal{B}}}\left[\mathcal{L}_{\text{A-OHL}}\left(\tilde{\boldsymbol{x}}\right)\right], \text { where } \mathcal{L}_{\text{A-OHL}}(\tilde{\boldsymbol{x}}) =\\
& \sum_{\tilde{\boldsymbol{x}} \in \tilde{\mathcal{B}}}-\frac{1}{|\tilde{\mathcal{B}}|}  \log \frac{\exp (\frac {z(\tilde{\boldsymbol{x}})^T z(\tilde{\boldsymbol{w}})} {\tau})}{\sum\limits_{\boldsymbol{h} \in \mathcal{B}_h} \exp (\frac {z(\tilde{\boldsymbol{x}})^T z(\boldsymbol{h})} {\hat{\tau}_{c(\boldsymbol{h})}}) + \exp (\frac {z(\tilde{\boldsymbol{x}})^T z(\tilde{\boldsymbol{w}})} {\tau})}.
\end{aligned}
\end{equation}
where $\tilde{\mathcal{B}}$ is a batch of informative outliers from $\mathcal{D}^{\text{train}}_{\text{out}}$, $\mathcal{B}_\text{h}$ is a set of head-class samples in $\mathcal{B}$, $\hat{\tau}_{c(\boldsymbol{h})}$ denotes a dynamic temperature parameter designed for a head class that is explicitly given in Eq.(\ref{eq4}), and $\tilde{\boldsymbol{w}}$ represents the outlier-class prototype tailored for outliers.
We define $\tilde{\boldsymbol{w}}$ as a trainable optimization parameter of the model and can be updated through gradient descent. Specifically, the weight associated with the outlier class (referred to as the ($C$+1)-th class) is obtained from the ($C$+1)-way classifier and then transformed through a projection head (MLP) to produce the outlier-class prototype, which enhancing the cohesion and compact representation of the outliers. 

The main difference between A-OHL and previous works \cite{pascl,EAT,COCL,TSCL} is the dynamic temperature parameter $\hat{\tau}_{c(\boldsymbol{x})}$, which is tailored for head classes and is utilized to adjust the temperature between a head-class sample and an OOD sample during training.

\subsection{Dynamic Class-wise Temperature Adjustment}

To fulfill the necessity of dynamic temperature adjustment in tail-specific and head-specific class learning in Section \ref{Separate Class Learning}, we propose  dynamic class-wise temperature adjustment, diverging from the conventional practice of employing a static temperature value across all sample pairs in previous works \cite{pascl,EAT,COCL,TSCL}. The formulation is as follows: 
\begin{equation}
\label{eq4}
\hat{\tau}_{c(\boldsymbol{x})}=\tau \cdot\left[1-\sqrt{\frac{e}{E}} \cdot \sqrt{\hat{n}_{c(\boldsymbol{x})}}\right], (c(\boldsymbol{x}) \in\{1, 2, \ldots, C\})
\end{equation} 
where $e$ denotes the current training epoch, $E$ indicates the total training epochs, $c(\boldsymbol{x})$ signifies the ground-truth label of sample $\boldsymbol{x}$, and $\tau$ is a static temperature value. 
We first construct a vector $\xi=\left(n_1, n_2, \ldots, n_C\right)$ based on the number of samples for each ID class, where $n_i (i \in\{1,2, \ldots, C\})$ represents the number of samples for the $i$-th class. For computational simplicity, we then subject $\xi$ to $\ell_2$ normalization to  obtain $\hat{\xi}=\left(\hat{n}_1, \hat{n}_2, \ldots, \hat{n}_C\right)$.

Notably, our proposed temperature adjustment is contingent on the training epoch and the distributions of class samples. During training, the adjusted temperature value $\hat{\tau}$ gradually decreases, intensifying the pulling or pushing effect between the anchor and positive/negative sample. Specifically:


1) Regarding tail-specific class learning, as illustrated in left diagram in Fig.\ref{fig2}(c), a tail-class sample is employed as an anchor, while other tail samples and tail-class prototypes assigned to each tail class are considered positive samples, and OOD samples are viewed as negative samples. Through the dynamic reduction of the temperature value $\hat{\tau}$ between the anchor and positive samples, we amplify the pulling effect within each tail class. This refinement augments the compactness of tail classes, rendering them more distinguishable from OOD samples.

2) For head-specific class learning, as illustrated in right diagram in Fig.\ref{fig2}(c), an outlier serves as an anchor, with additional outliers and the outlier-class prototype acting as positive samples, while head-class samples are treated as negative samples. By dynamically decreasing the temperature value $\hat{\tau}$ between the anchor and negative samples, we intensify the pushing effect between OOD samples and head classes, thereby enhancing their distinguishability.

Through enhancing the feature representation of the tail class and reducing the overconfidence of the head class, the model can better distinguish OOD samples from both head and tail classes, thus improving the overall performance. 
Further discussion is provided in Appendix \ref{More Discussion about Dynamic Class-wise Temperature Adjustment}. 

\subsection{Informative Outlier Mining}
\label{Informative Outlier Mining}
Instead of randomly sampling outliers from the auxiliary OOD training data \cite{OE,energy,OECC,SOFL,pascl,EAT,TSCL,COCL}, we propose informative outlier mining to improve the quality of auxiliary outliers. The aim of informative outlier mining is to identify and utilize diverse types of outliers based on their affinity with head and tail classes. Each candidate outlier is evaluated through an outlier score function formulated as follows:
\begin{equation}
\label{eq5}
\mathcal{S(\tilde{\boldsymbol{x}})}= 1- \sum\nolimits_{n=1}^N\log \frac{\exp(f_n(\tilde{\boldsymbol{x}}))}{\sum_{j=1}^C \exp(f_j(\tilde{\boldsymbol{x}}))},
\end{equation}
where $\tilde{\boldsymbol{x}}$ is an auxiliary outlier, $N$ represents the number of head classes, $f_n(\tilde{\boldsymbol{x}})$ indicates the $n$-th index of head class of $f(\tilde{\boldsymbol{x}})$. 

In Eq.(\ref{eq5}), we use the sum of softmax probabilities calculated over the ID tail classes as the outlier score, which serves as a measure of the likelihood of an outlier being classified as a tail or head class, or neither. Algorithm \ref{algorithm1} in Appendix \ref{The Informative Outlier Mining Algorithm} provides the complete process of informative outlier mining. 

As illustrated in Figure \ref{fig2}(b), after the outlier mining process, outliers are evenly classified into three distinct categories:
1) $\emph{Tail-class-like OOD}$: Outliers that are closer in feature space to tail samples and are thus more likely to be misclassified as tail classes.
2) $\emph{Head-class-like OOD}$: Outliers that exhibit higher similarity to head samples, making them prone to being predicted as head classes.
3) $\emph{Neutral OOD}$: Outliers with no clear similarity to head or tail classes, equally likely to be misclassified as either.
Further details can be found in Appendix \ref{The Informative Outlier Mining Algorithm}.



\subsection{Training}
\label{4.4}

The overall training objective of our approach is given as follows: 
\begin{equation}
\label{eq6}
\mathcal{L}_{\text {RSCL }}=\mathcal{L}_{\text {OCL}}+\beta \cdot \mathcal{L}_{\text {tail }}+\gamma \cdot \mathcal{L}_{\text {head }},
\end{equation}
where $\mathcal{L}_{\text {OCL}}$ is defined in Eq.(\ref{eq1}), $\mathcal{L}_{\text {tail}}$  represents the A-TSCL defined in Eq.(\ref{eq2}), and $\mathcal{L}_{\text {head}}$ denotes the A-OHL defined Eq.(\ref{eq3}). Hyper-parameters $\beta$  and $\gamma$ regulate the weighting of $\mathcal{L}_{\text {tail}}$ and $\mathcal{L}_{\text {head}}$.


The complete RSCL training process is detailed in  Algorithm \ref{algorithm2} in Appendix \ref{The RSCL Algorithm}.
Specifically, during the first $\frac{3}{4}$ of the total training epochs, a subset of mixed outliers drawn from the three distinct OOD categories is utilized. By incorporating diverse outliers, the model is exposed to a broad spectrum of OOD features, thus facilitating the acquisition of comprehensive OOD knowledge. Subsequently, in the final $\frac{1}{4}$ of the total training epochs, outliers exclusively from the $\emph{Neutral OOD}$ category are utilized. 
These outliers are more ambiguous and challenging, and focusing on them in later training stages helps refine the model’s OOD discrimination capability.
Further discussion on the advantage of utilizing diverse types of outliers at distinct training epochs is provided in Appendix \ref{More Ablation Study}. 

\subsection{Inference}
\label{4.5}
At inference time, a simple thresholding mechanism is used for detecting OOD samples. In particular, the OOD detector $G_\eta(\boldsymbol{x})$ can be constructed using the ($C$+1)-th class as follows:
\begin{equation}
\label{eq7}
G_\eta(\boldsymbol{x})= \begin{cases}1 & \text { if } F_{C+1}(\boldsymbol{x}) \geq \eta \\ 0 & \text { if } F_{C+1}(\boldsymbol{x})<\eta,\end{cases}
\end{equation}
where $F_{C+1}(\boldsymbol{x})$ denotes the softmax score corresponding to the outlier class for sample $\boldsymbol{x}$. The threshold $\eta$ is determined by choosing an appropriate value to ensure that a significant proportion (e.g., \textbf{95$\%$}) of the OOD test data is correctly classified, following \cite{pascl, EAT, TSCL}. A sample $\boldsymbol{x}$ is classified as OOD if the OOD score $F_{C+1}(\boldsymbol{x})$ is above the threshold $\eta$, and vice versa.

\begin{table}[ht]
\centering
\caption{OOD Detection Performance of RSCL, OE, and PASCL on CIFAR10/100-LT  using ResNet-18. ``Average'' means the results averaged across six $\mathcal{D}_{\text {out}}^{\text {test}}$ sets. Bold numbers are the best results.}
\label{table1}
\resizebox{\linewidth}{!}{  
\begin{tabular}{ccccc}
\hline
\multirow{2}{*}{$\mathcal{D}_{\text {in}}^{\text {train}}$} & \multirow{2}{*}{$\mathcal{D}_{\text {out}}^{\text {test}}$}  & \textbf{AUROC$\uparrow$}   & \textbf{AUPR$\uparrow$}  & \textbf{FPR95$\downarrow$}       \\ \cline{3-5} 
                    &                      & \multicolumn{3}{c}{OE \cite{OE} / PASCL \cite{pascl}  / \textbf{RSCL (Ours)}} \\ \hline
\multirow{6}{*}{\begin{tabular}[c]{@{}c@{}}CIFAR10-LT
\end{tabular}}      & Texture   & 90.44 / 92.74 / \textbf{95.86}  & 74.04 / 84.08 / \textbf{93.58}    & 22.90 / 22.36 / \textbf{21.80} \\
                    & SVHN   & 92.39 / 95.54 / \textbf{98.43}  & 93.27 / 97.37 / \textbf{99.35}   & 16.14 / 14.01 / \textbf{07.59}   \\
                    & CIFAR100    &83.15 / 84.00
                    / \textbf{86.51}  & 78.22 / 81.90 / \textbf{86.67}   & 55.42 / 57.04 / \textbf{51.67}    \\
                    & Tiny ImageNet       & 85.58 / 86.71 / \textbf{89.78}  & 75.39 / 82.23 / \textbf{86.62}   & 44.71 / 50.21 / \textbf{40.25}    \\
                    & LSUN      & 90.88 / 92.93 / \textbf{95.99}  & 85.71 / 91.17 / \textbf{95.70}   & 26.00 / 26.69 / \textbf{18.85}    \\
                    & Places365 & 89.25 / 91.17 / \textbf{94.19}  & 93.95 / 96.02 / \textbf{97.76}   & 31.70 / 31.99 / \textbf{26.25}   \\
                    \cline{2-5} & \textbf{Average}  & 88.61 / 90.53 / \textbf{93.46}  & 83.43 / 88.49 / \textbf{93.28}   & 32.81 / 32.99 / \textbf{27.73}   \\ 
\hline
\multirow{6}{*}{\begin{tabular}[c]{@{}c@{}}CIFAR100-LT
\end{tabular}}      & Texture & 76.07 / 76.16 / \textbf{76.79}  & 58.00 / 58.02 / \textbf{64.60}   & 67.53 / \textbf{66.27} / 68.81     \\
                    & SVHN    & 78.53 / 78.72 / \textbf{85.31}  & 87.42 / 87.12 / \textbf{90.93}   & 58.19 / 54.78 / \textbf{42.46}     \\
                    & CIFAR10 & 62.07 / 62.35 / \textbf{63.14}  & 56.79 / \textbf{57.17} / 56.87   & 79.87 / 79.88 / \textbf{76.98}    \\
                    & Tiny ImageNet & 68.15 / 68.41 / \textbf{69.96}  & 51.57 / 52.09 / \textbf{54.52}   & 75.97 / 76.06 / \textbf{73.55}    \\
                    & LSUN     & 76.80 / 77.11 / \textbf{81.78}  & 60.50 / 61.47 / \textbf{68.49}   & 64.38 / 63.72 / \textbf{55.62}   \\
                    & Places365  & 75.49 / 75.85 / \textbf{79.68}  & 86.10 / 86.53 / \textbf{89.15}   & 65.30 / 65.50 / \textbf{59.73}   \\
                    \cline{2-5} & \textbf{Average}  & 72.85 / 73.10 / \textbf{76.11}  & 66.73 / 67.06 / \textbf{70.76}   & 68.54 / 67.70 / \textbf{62.86}   \\
\hline
\end{tabular}
}

\end{table}

\begin{table}[]
\centering
\footnotesize 
\caption{Classification accuracy on head and tail classes of RSCL, OE, and PASCL.} 
\label{table2}
\resizebox{\linewidth}{!}{  
\begin{tabular}{c|ccc|ccc|ccc}
\hline
\multirow{2}{*}{Methods} & \multicolumn{3}{c|}{$\mathcal{D}_{\text {in}}^{\text {train}}$: CIFAR10-LT}      & \multicolumn{3}{c|}{$\mathcal{D}_{\text {in}}^{\text {train}}$: CIFAR100-LT}     & \multicolumn{3}{l}{$\mathcal{D}_{\text {in}}^{\text {train}}$: ImageNet-LT}      \\ \cline{2-10} 
                         & Head           & Tail           & Overall        & Head           & Tail           & Overall        & Head           & Tail           & Overall        \\ \hline
OE                       & 86.03          & 60.93          & 73.48          & 60.66          & 19.32          & 39.99          & 59.12          & 27.28          & 43.20          \\
PASCL                    & 86.44          & 65.44          & 75.94          & 60.55          & 19.35          & 39.95          & 60.98          & 29.51          & 45.29          \\
\textbf{RSCL(ours)}      & \textbf{87.53} & \textbf{70.44} & \textbf{79.14} & \textbf{62.76} & \textbf{22.86} & \textbf{42.81} & \textbf{61.57} & \textbf{30.66} & \textbf{46.12} \\ \hline
\end{tabular}
}
\end{table}
\section{Experiments and Results}
\subsection{Experiment Settings}
\label{Experiment Settings}

\textbf{Datasets}.
\quad We use three popular long-tailed image classification datasets as ID training data (i.e., $\mathcal{D}_{\text {in}}^{\text {train}}$), including CIFAR10-LT, CIFAR100-LT \cite{cao2019learning}, and ImageNet-LT \cite{liu2019large}.  
For CIFAR10-LT and CIFAR100-LT, the imbalance ratio $\rho$ is set to 100 \cite{LA,pascl}.
The original CIFAR test sets and the ImageNet validation set serve as ID test data ($\mathcal{D}_{\text{in}}^{\text{test}}$).
Following \cite{OE,energy,pascl,EAT}, 300k Random Images is used as OOD training data ($\mathcal{D}_{\text {out}}^{\text {train}}$) for CIFAR10/100-LT and ImageNet-Extra \cite{pascl} is used as $\mathcal{D}_{\text {out}}^{\text {train}}$ for ImageNet-LT.  
For OOD test set ($\mathcal{D}_{\text {out}}^{\text {test}}$),
we use six datasets including Texture \cite{texture}, SVHN \cite{SVHN}, CIFAR \cite{CIFAR}, Tiny ImageNet \cite{Tinyimagenet}, LSUN \cite{LSUN}, and Places365 \cite{Places365} introduced in the SC-OOD benchmark \cite{yang2021semantically} for CIFAR10/100-LT. We use ImageNet-1k-OOD constructed by \cite{pascl} as $\mathcal{D}_{\text {out}}^{\text {test}}$ to ImageNet-LT. 
More details about datasets are presented in Appendix \ref{appendix_datasets}.

\textbf{Evaluation Metrics}. \quad 
Following \cite{pascl, EAT, TSCL}, we evaluate our approach using the below metrics:
(1) \textbf{AUROC}: The area under the receiver operating characteristic curve. (2) \textbf{AUPR}: The area under the precision-recall curve, where we treat OOD samples as positive. (3) \textbf{FPR95}: The false positive rate (FPR) when the true positive rate (TPR) of OOD samples is at 95$\%$. (4) \textbf{ACC}: The accuracy achieved on the entire ID test set.

\textbf{Implementation Details}.
\quad We compare our RSCL with several existing OOD detection methods on long-tailed training sets, including classical methods MSP \cite{MSP}, OE \cite{OE}, EnergyOE \cite{energy}, DAC \cite{DAC}, SOFL \cite{SOFL}, NTOM \cite{atom}, and very recently published methods PASCL \cite{pascl}, EAT \cite{EAT}, COCL \cite{COCL}, and TSCL \cite{TSCL}. For experiments on CIFAR10-LT and CIFAR100-LT, we use the standard ResNet18 \cite{Resnet} as the backbone. For experiments on ImageNet-LT, we follow the settings in \cite{pascl,EAT,COCL} and use ResNet50 \cite{Resnet} as the backbone. Appendix \ref{Implementation Details} provides more implementation details. 

\subsection{Main Results}
\label{Main Results and Discussion}
\textbf{Comparison with the baseline methods OE \cite{OE} and PASCL\cite{pascl}}. \quad Table \ref{table1} presents a comprehensive comparison between our RSCL and two baseline methods, OE and PASCL. OE is a representative OOD detection technique based on class-balanced training sets, while PASCL is a pioneering work for long-tailed OOD detection. The results illustrate that our RSCL significantly improves the OOD detection performance on each OOD test set compared to OE and PASCL. For example, on the CIFAR10-LT model, compared to PASCL, our RSCL improves the average AUROC by 2.93$\%$, reduces the average FPR95 by 5.26$\%$, and enhances the average AUPR by 4.79$\%$.

\textbf{Improvements on head and tail in-distribution classes.} 
\quad Table \ref{table2}  shows the classification accuracy of our RSCL over OE and PASCL on head and tail classes. Notably, our RSCL demonstrates substantial enhancements in classification accuracy across both head and tail classes, thereby benefiting the overall ID performance. Compared to PASCL\cite{pascl}, known for its pronounced bias towards tail classes with limited improvements in head classes, our RSCL achieves a good balance.

\textbf{Comparison with other competitive methods.}
\quad Table \ref{table3} and Table \ref{table4} report the results for the CIFAR-LT and ImageNet-LT datasets, respectively. The results reveal that when the training set follows a long-tail distribution, existing post-hoc OOD detection methods \cite{MSP} suffer a notable decline in their ability to detect OOD samples. Furthermore, methods that utilize auxiliary outlier data \cite{DAC, energy, SOFL, atom} exhibit significant potential for improving OOD detection, but their performance on ID classification remains relatively poor. This is due to the absence of effective outlier mining that account for the affinity between outliers and both head and tail classes. Consequently, the model struggles to accurately differentiate between ID and OOD samples, resulting in misclassifications of ID samples. In particular, our RSCL also outperforms recently proposed methods \cite{pascl,EAT, TSCL, COCL}. This consistent improvement on both ID and OOD data is due to our dynamic temperature adjustment and informative outlier mining, which enhances the discrimination ability of OOD samples from both head and tail classes.



\begin{table}[]
\centering
\caption{Comparison with other methods on CIFAR10-LT and CIFAR100-LT using ResNet18.  Mean and standard deviation over six random runs are reported. Bold numbers are the best results.} 
\label{table3}
\resizebox{\linewidth}{!}{  
\begin{tabular}{c|cccc|cccc}
\hline
\multirow{2}{*}{Method} & \multicolumn{4}{c|}{$\mathcal{D}_{\text {in}}^{\text {train}}$: CIFAR10-LT}                       & \multicolumn{4}{c}{$\mathcal{D}_{\text {in}}^{\text {train}}$: CIFAR100-LT}                       \\ \cline{2-9} 
                         & AUROC↑         & AUPR↑          & FPR95↓         & ACC↑           & AUROC↑         & AUPR↑          & FPR95↓         & ACC↑           \\ \hline
MSP \cite{MSP}                      & 68.68          & 67.69          & 75.20          & 69.84          & 61.38          & 58.22          & 83.15          & 41.14          \\
EnergyOE \cite{energy}                 &  $87.98_{\pm 0.46}$          &  $84.55_{\pm 1.24}$          & $37.71_{\pm 0.51}$          & $66.53_{\pm 1.31}$          & $68.76_{\pm 0.86}$          & $62.15_{\pm 0.47}$          & $72.87_{\pm 1.29}$          & $33.87_{\pm 0.37}$          \\
DAC \cite{DAC}                     & $89.91_{\pm 0.45}$           & $86.26_{\pm 1.27}$           & $31.99_{\pm 0.64}$           & $75.30_{\pm 0.70}$           & $72.51_{\pm 0.30}$           & $66.60_{\pm 0.40}$           & $68.96_{\pm 0.70}$           & $41.54_{\pm 0.60}$           \\
SOFL \cite{SOFL}                    & $90.60_{\pm 0.56}$          & $89.26_{\pm 1.44}$          & $33.67_{\pm 0.86}$          & $70.59_{\pm 1.14}$          & $72.97_{\pm 0.78}$          & $67.99_{\pm 1.08}$          & $68.73_{\pm 0.80}$          & $40.55_{\pm 0.84}$          \\
NTOM \cite{atom}                    & $90.61_{\pm 0.51}$          & $90.19_{\pm 1.06}$          & $36.53_{\pm 0.77}$          & $68.78_{\pm 0.81}$          & $70.90_{\pm 0.64}$          & $66.76_{\pm 0.79}$          & $72.59_{\pm 0.67}$          & $35.68_{\pm 1.02}$          \\
EAT \cite{EAT}                     & $90.98_{\pm 0.28}$          & $89.34_{\pm 0.54}$          & $33.01_{\pm 0.67}$          & $78.74_{\pm 0.37}$          & $72.97_{\pm 0.53}$          & $67.75_{\pm 0.45}$          & $67.91_{\pm 0.80}$          & $42.71_{\pm 0.63}$          \\
COCL \cite{COCL}                     & $91.66_{\pm 0.33}$          & $91.06_{\pm 0.46}$          & $34.40_{\pm 0.58}$          & $74.87_{\pm 0.63}$          & $74.68_{\pm 0.65}$          & $\textbf{70.79}_{\pm 0.38}$ & $67.87_{\pm 0.79}$           & $41.80_{\pm 0.58}$           \\
TSCL\cite{TSCL}                     & $92.06_{\pm 0.08}$          & $91.01_{\pm 0.25}$          & $30.19_{\pm 0.36}$          & $76.10_{\pm 0.51}$         & $74.08_{\pm 0.30}$          & $68.10_{\pm 0.21}$          & $65.36_{\pm 0.68}$          & $40.11_{\pm 0.26}$          \\ \hline
\textbf{RSCL(ours)}      & $\textbf{93.46}_{\pm 0.06}$ & $\textbf{93.28}_{\pm 0.09}$ & $\textbf{27.73}_{\pm 0.25}$ & $\textbf{79.14}_{\pm 0.34}$ & $\textbf{76.11}_{\pm 0.34}$ & $70.76_{\pm 0.33}$          & $\textbf{62.86}_{\pm 0.58}$ & $\textbf{42.81}_{\pm 0.40}$ \\ \hline
\end{tabular}
}

\end {table}

\begin{table}[]
\centering
\footnotesize 
\caption{Results on ImageNet-LT using ResNet50. Bold numbers are the best results.} 
\label{table4}
\begin{tabular}{cccccc}
\hline
$\mathcal{D}_{\text {in}}^{\text {train}}$     & Method      & AUROC↑ & AUPR↑ & FPR95↓ & ACC↑ \\ \hline
\multirow{11}{*}{\begin{tabular}[c]{@{}c@{}}ImageNet-LT
\end{tabular}} 
        & MSP\cite{MSP}         & 53.81           & 51.63           & 90.15          & 39.65   \\
        & OE \cite{OE}  & 68.36           & 70.46           & 87.31          & 43.20   \\
        & EnergyOE \cite{energy}  & 75.57           & 74.28           & 78.18          & 35.67   \\
        
        & DAC \cite{DAC}         & 75.61           & 74.06           & 75.49          & 39.89   \\
        & SOFL \cite{SOFL}        & 76.18          & 74.65           & 74.03          & 38.78   \\
        & NTOM \cite{atom}         & -           & -           & -          & -   \\
        & PASCL \cite{pascl}      & 68.63           & 70.64           & 86.78          & 45.29   \\
        
        & EAT \cite{EAT}  & 69.55  & 69.10  & 86.66 & 43.74 \\
        & COCL \cite{COCL}  & 72.10  & 72.61  & 86.11 & 42.71 \\
        & TSCL \cite{TSCL} & 75.04  & 73.98  & 78.65 & 43.89 \\ \hline
        & \textbf{RSCL (ours)} & \textbf{78.90}  & \textbf{78.10}  & \textbf{73.04} & \textbf{46.19} \\ \hline
\end{tabular}
\vspace{-3mm}
\end {table}

\begin{figure}[ht] 
	\centering  
	\subfigure[$\mathcal{L}_\text{OCL}$ Loss]{
		\label{fig3a}
		\includegraphics[width=0.47\linewidth]{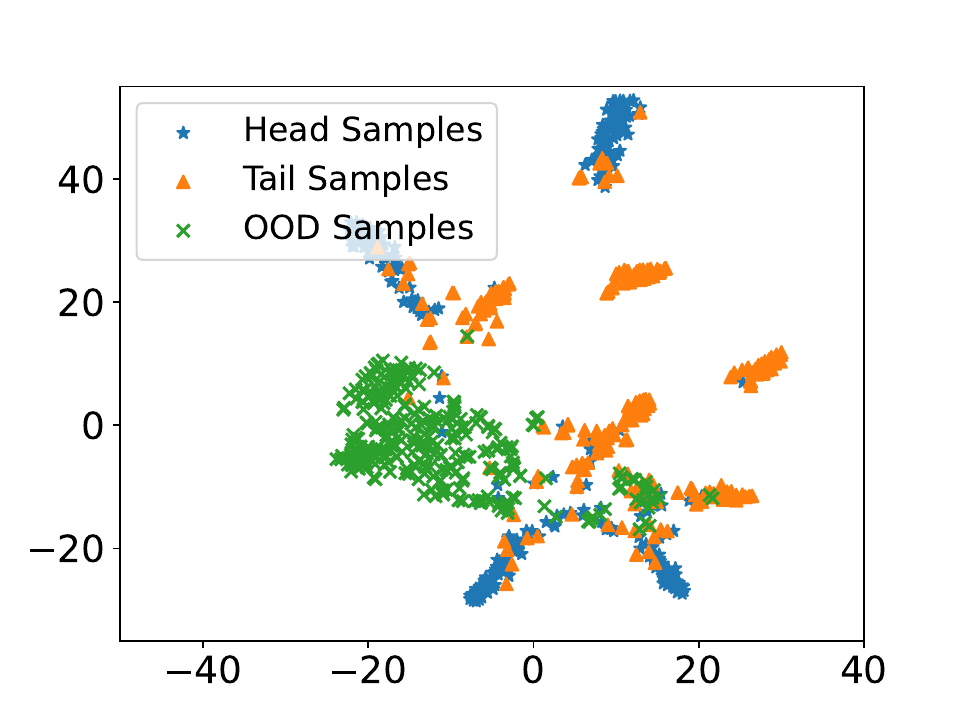}}
	\subfigure[RSCL Loss (ours)]{
		\label{fig3b}
		\includegraphics[width=0.47\linewidth]{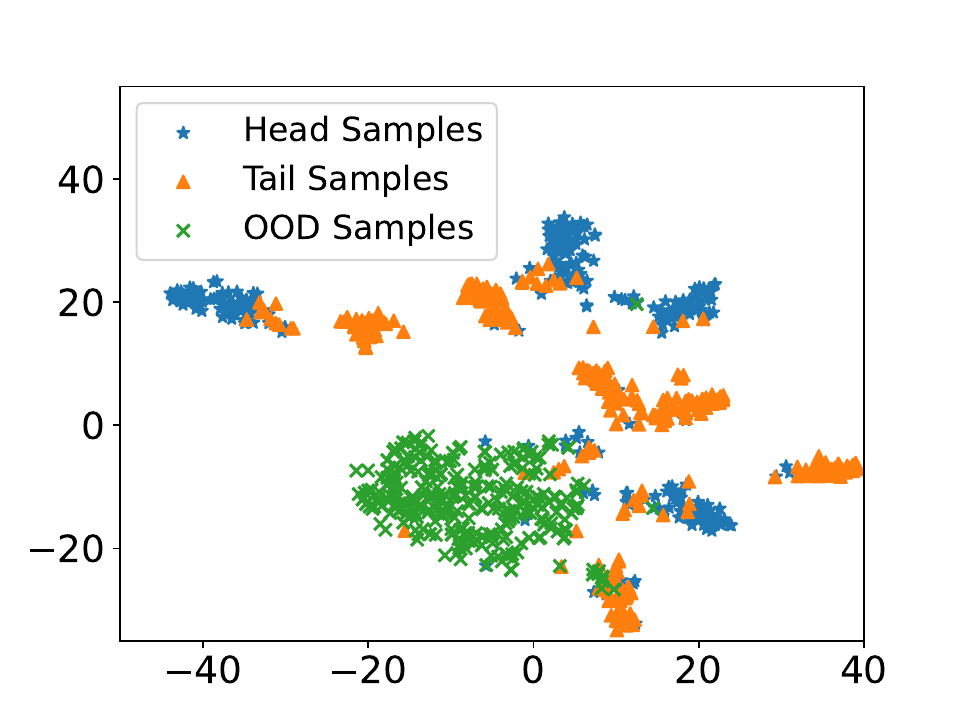}}
        \caption{t-SNE \citep{tsne} visualizations of penultimate-layer features from ResNet-18 trained on CIFAR10-LT using two different losses: (a) $\mathcal{L}_\text{OCL}$ loss as defined in Eq.(\ref{eq1}), and (b) Our loss as defined in Eq.(\ref{eq6}). We randomly select 300 head samples (75 per head class), 300 tail samples (50 per tail class), and 300 OOD samples from SVHN. The default value of $k$ is set to 0.6.}
	\label{fig3}
\end{figure}

\textbf{Mitigating the confusion between OOD samples and head/tail-class samples.}
\quad Figure \ref{fig3} presents a visualization of the feature embeddings corresponding to tail-class samples, head-class samples, and OOD samples. The feature distributions depicted in Figure \ref{fig3a} reveal that  OOD samples are easily confused with some tail samples and a few head samples in long-tailed scenarios. However, when training the model with our RSCL objective in Eq.(\ref{eq6}), a discernible enhancement in discriminating between OOD samples and head/tail samples is evident, as illustrated in Figure \ref{fig3b}. This observation highlights the efficacy of our RSCL for long-tailed OOD detection.

\subsection{Ablation Studies}
\label{Ablation Studies}
In this section, we provide the ablation results to analyze the influence of various factors in our approach. For consistency, our analysis is focused on the  CIFAR100-LT benchmark. However, similar trends are observed on the CIFAR10-LT and ImageNet-LT benchmark as well.

\begin{table}[h]
\centering
\tiny
\caption{The effectiveness of different loss components. All values are percentages averaged over six OOD test datasets. ${\mathcal{L}_{\text{tail}}}^{\star}$ and ${\mathcal{L}_{\text{head}}}^{\star}$ represent the vanilla loss without incorporating the informative outlier mining (OM) and dynamic class-wise temperature adjustment (TA). }
\label{table5}
\resizebox{\linewidth}{!}{  
\begin{tabular}{ccccccccc}
\hline
\multicolumn{5}{c}{Loss Components} & \multirow{2}{*}{AUROC↑} & \multirow{2}{*}{AUPR↑} & \multirow{2}{*}{FPR95↓} & \multirow{2}{*}{ACC↓} \\ \cline{1-5}
$\mathcal{L}_{\text{OCL}}$    & ${\mathcal{L}_{\text{tail}}}^{\star}$
& ${\mathcal{L}_{\text{head}}}^{\star}$   &$\text{OM}$ & $\text{TA}$  &  &  &  &     \\ \hline
 $\checkmark$ &  &   &  &  & 73.93  & 67.13  & 64.90  & 39.94             \\ \hline
$\checkmark$    & $\checkmark$ &$\checkmark$  &    &    & 74.15  & 68.34  & 65.81  & 40.11    \\ \hline
$\checkmark$ & $\checkmark$  & $\checkmark$ & $\checkmark$ &  & 75.29  & 69.25  & 63.33  & 41.70             \\ \hline
$\checkmark$ & $\checkmark$  & $\checkmark$ &  & $\checkmark$ & 75.23  & 70.29  & 64.88  & 42.05             \\ \hline

$\checkmark$    & $\checkmark$   &  &$\checkmark$ & $\checkmark$ & 74.87  & 68.09 & 63.63  & 40.25 \\ \hline
$\checkmark$    &    & $\checkmark$ &$\checkmark$ & $\checkmark$ & 75.06  & 69.72 & 64.57  & \textbf{43.02} \\ \hline
$\checkmark$    &  $\checkmark$  & $\checkmark$ &$\checkmark$ & $\checkmark$ & \textbf{76.11}  & \textbf{70.76} & \textbf{62.86}  & 42.81 
 \\ \hline
\end{tabular}
 }
\end{table}

\noindent \textbf{The effectiveness of different loss components.} 
\quad Our RSCL consists of three different loss components: (1) $\mathcal{L}_{\text{OCL}}$, (2) $\mathcal{L}_{\text{tail}}$ = (${\mathcal{L}_{\text{tail}}}^{\star}$ + OM + TA), and (3) $\mathcal{L}_{\text{head}}$ = (${\mathcal{L}_{\text{head}}}^{\star}$ + OM + TA). The ablation results of these three loss components are shown in Table \ref{table5}. It is evident that the use of $\mathcal{L}_{\text{OCL}}$ in combination with either $\mathcal{L}_{\text{tail}}$ or $\mathcal{L}_{\text{head}}$ leads to marginal enhancements in OOD detection and ID classification. Optimal model performance is achieved when all three components are utilized simultaneously. This verifies the effectiveness of $\mathcal{L}_{\text{tail}}$ and $\mathcal{L}_{\text{head}}$ in improving model performance.

\textbf{The effectiveness of informative outlier mining.}  
\quad Informative outlier mining plays a key role in our approach. As shown in Table \ref{table5} (line 2 and line 3), once incorporating this strategy, notable improvements are observed: the AUROC improves from 74.15$\%$ to 75.29$\%$, the FPR95 reduces from 65.81$\%$ to 63.33$\%$, and the ID classification accuracy enhances from 40.11$\%$ to 41.70$\%$. This suggests its effectiveness in enhancing both OOD detection  and ID classification performance.

\textbf{The effectiveness of dynamic class-wise temperature adjustment.}
\quad Once incorporating the temperature adjustment strategy, significant enhancements are observed from Table \ref{table5} (line 2 and line 4): the AUROC improves from 74.15$\%$ to 75.23$\%$, the FPR95 decreases from 65.81$\%$ to 64.88$\%$, and the ID classification accuracy enhances from 40.11$\%$ to 42.05$\%$. This demonstrates the efficacy of this adjustment in improving both the OOD detection performance and ID classification accuracy.

\textbf{More Ablation Experiments.} Additional ablation studies are provided in Appendix \ref{More Ablation Study}.
\section{Conclusion}
This paper introduces Refined Separate Class Learning (RSCL), a novel approach designed to effectively distinguish OOD samples from both head and tail classes.
RSCL addresses the constraints present in the original SCL approach by incorporating dynamic class-wise temperature adjustment and informative outlier mining. 
Specifically, dynamic class-wise temperature adjustment is proposed to modulate the temperature value for each head and tail class, while informative outlier mining is introduced to identify diverse types of outliers based on their affinity with head and tail classes. Comprehensive evaluation and ablation studies demonstrate that RSCL significantly enhances the OOD detection performance while improving the classification accuracy on in-distribution data.







\clearpage
\bibliography{references.bib}
\bibliographystyle{plainnat}

\appendix
\newpage

\section{More Discussion about Dynamic Class-wise Temperature Adjustment}
\label{More Discussion about Dynamic Class-wise Temperature Adjustment}


\subsection{Class Distributions}
To better illustrate the class distribution patterns, we use CIFAR10 and CIFAR100 \cite{CIFAR} as representative examples, given the large number of classes (1,000) in ImageNet-1k \cite{imagenet}.


For the CIFAR10 training set with an imbalance ratio of $\rho = 100$, the number of samples per in-distribution (ID) class is represented as $\xi=\left(n_1, n_2, \ldots, n_{10}\right)$ = (5000, 2997, 1796, 1077, 645, 387, 232, 139, 83, 50). 
Applying $\ell_2$ normalization to $\xi$ yields the normalized class distribution $\hat{\xi}=\left(\hat{n}_1, \hat{n}_2, \ldots, \hat{n}_{10}\right)$ = (0.8005, 0.4798, 0.2875, 0.1724, 0.1033, 0.0620, 0.0371, 0.0223, 0.0133, 0.0080).

For the CIFAR100 training set with an imbalance ratio of $\rho = 100$, the number of samples per in-distribution (ID) class is represented as $\xi=\left(n_1, n_2, \ldots, n_{100}\right)$ = (500, 477, 455, 434, 415, 396, 378, 361, 344, 328, 314, 299, 286, 273, 260, 248, 237, 226, 216, 206, 197, 188, 179, 171, 163, 156, 149, 142, 135, 129, 123, 118, 112, 107, 102, 98, 93, 89, 85, 81, 77, 74, 70, 67, 64, 61, 58, 56, 53, 51, 48, 46, 44, 42, 40, 38, 36, 35, 33, 32, 30, 29, 27, 26, 25, 24, 23, 22, 21, 20, 19, 18, 17, 16, 15, 15, 14, 13, 13, 12, 12, 11, 11, 10, 10, 9, 9, 8, 8, 7, 7, 7, 6, 6, 6, 6, 5, 5, 5, 5). After applying $\ell_2$ normalization to $\xi$, the normalized class distribution $\hat{\xi}=\left(\hat{n}_1, \hat{n}_2, \ldots, \hat{n}_{100}\right)$ = (0.2986, 0.2849, 0.2717, 0.2592, 0.2478, 0.2365, 0.2257, 0.2156, 0.2054, 0.1959, 0.1875, 0.1786, 0.1708, 0.1630, 0.1553, 0.1481, 0.1415, 0.1350, 0.1290, 0.1230, 0.1176, 0.1123, 0.1069, 0.1021, 0.0973, 0.0932, 0.0890, 0.0848, 0.0806, 0.0770, 0.0735, 0.0705, 0.0669, 0.0639, 0.0609, 0.0585, 0.0555, 0.0532, 0.0508, 0.0484, 0.0460, 0.0442, 0.0418, 0.0400, 0.0382, 0.0364, 0.0346, 0.0334, 0.0317, 0.0305, 0.0287, 0.0275, 0.0263, 0.0251, 0.0239, 0.0227, 0.0215, 0.0209, 0.0197, 0.0191, 0.0179, 0.0173, 0.0161, 0.0155, 0.0149, 0.0143, 0.0137, 0.0131, 0.0125, 0.0119, 0.0113, 0.0107, 0.0102, 0.0096, 0.0090, 0.0090, 0.0084, 0.0078, 0.0078, 0.0072, 0.0072, 0.0066, 0.0066, 0.0060, 0.0060, 0.0054, 0.0054, 0.0048, 0.0048, 0.0042, 0.0042, 0.0042, 0.0036, 0.0036, 0.0036, 0.0036, 0.0030, 0.0030, 0.0030, 0.0030).

\begin{table}[ht]
\small
\centering
\caption{Impact of linear vs. square root functions in dynamic class-wise temperature adjustment on OOD detection peformance and ID classification accuracy.} 
\label{table6}
\begin{tabular}{cccccc}
\hline
$\mathcal{D}_{\text {in}}^{\text {train}}$  & Modulation function      & \textbf{AUROC$\uparrow$} & \textbf{AUPR$\uparrow$} & \textbf{FPR95$\downarrow$} & \textbf{ACC$\uparrow$} \\ \hline
\multirow{3}{*}{\begin{tabular}[c]{@{}c@{}}CIFAR10-LT
\end{tabular}} 
        & Linear Function   & 92.15 & 91.71  & 32.23  & 76.77  \\
        & Square Root Function & \textbf{93.46} & \textbf{93.28}  & \textbf{27.73}  & \textbf{79.14}\\ \hline
\multirow{3}{*}{\begin{tabular}[c]{@{}c@{}}CIFAR100-LT
\end{tabular}} 
        & Linear Function   & 75.05 & 68.66 & 64.53 & 41.72   \\
        & Square Root Function  & \textbf{76.11} & \textbf{70.76} & \textbf{62.86} & \textbf{42.81} \\ \hline
\multirow{3}{*}{\begin{tabular}[c]{@{}c@{}}ImageNet-LT
\end{tabular}} 
        & Linear Function  & 78.62 & 77.95 & 74.28 & 45.80   \\
        & Square Root Function  & \textbf{78.90} & \textbf{78.10} & \textbf{73.04} & \textbf{46.19} \\ \hline
\end{tabular}
\end {table}

\subsection{The design of Dynamic Class-wise Temperature Adjustment}
\textbf{Incorporating Training Epochs and Class Sample Sizes}.
\quad In long-tailed scenarios, head classes (with abundant samples) and tail classes (with fewer samples) exhibit differing learning dynamics. To address this, our adjustment mechanism considers both the current training epoch ($e$) and the total training epochs ($E$), ensuring that temperature modulation evolves as training progresses. Additionally, the normalized class sample size, denoted as $\hat{n}_{c(\boldsymbol{x})}$, allows the temperature to adapt based on the relative abundance of each class.

\textbf{Rationale for the Square Root Function}.
\quad The inclusion of the square root function serves to moderate the influence of class sample size on temperature adjustment. In long-tailed distributions, the disparity between head and tail class sample sizes can be substantial. A linear adjustment might overemphasize this difference, leading to overly aggressive temperature scaling for tail classes. By applying the square root, we achieve a more balanced modulation, ensuring that temperature adjustments are neither too drastic for tail classes nor too minimal for head classes. 
As illustrated in Table \ref{table6}, compared with linear functions ($\hat{\tau}_{c(\boldsymbol{x})}=\tau \cdot\left[1-{\frac{e}{E}} \cdot {\hat{n}_{c(\boldsymbol{x})}}\right]$), using the square root function (in Eq.(\ref{eq4})) is more beneficial for OOD detection and ID classification.
In practice, this approach aligns with practices in statistical modeling where square root transformations are employed to stabilize variance and normalize distributions \cite{transformations,tukey1977exploratory,kutner2005applied}.

\begin{figure}[h] 
	\centering  
	\subfigure[Temperature variation of head class $n_1$]{
		\label{fig4a}
		\includegraphics[width=0.48\linewidth]{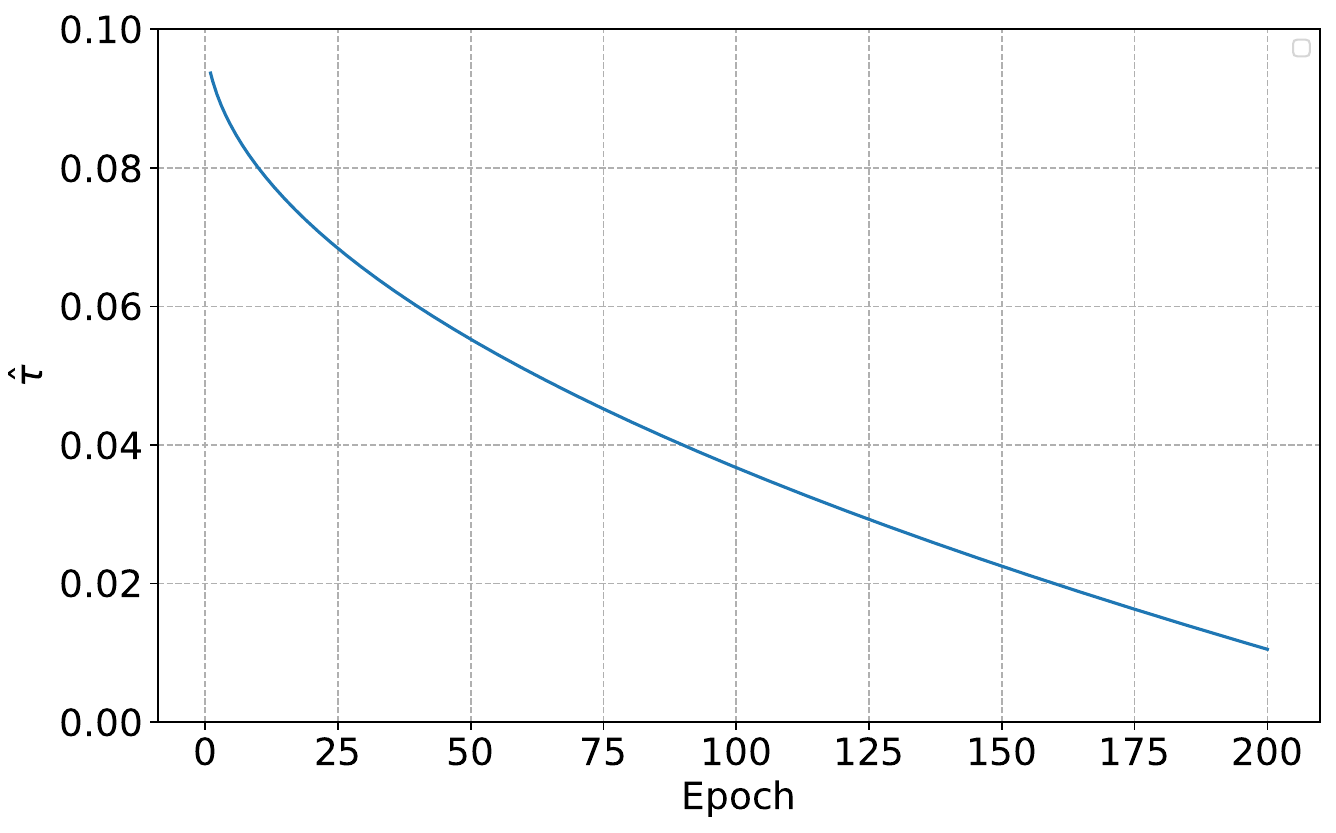}}
	\subfigure[Temperature variation of tail class $n_{5}$]{
		\label{fig4b}
		\includegraphics[width=0.48\linewidth]{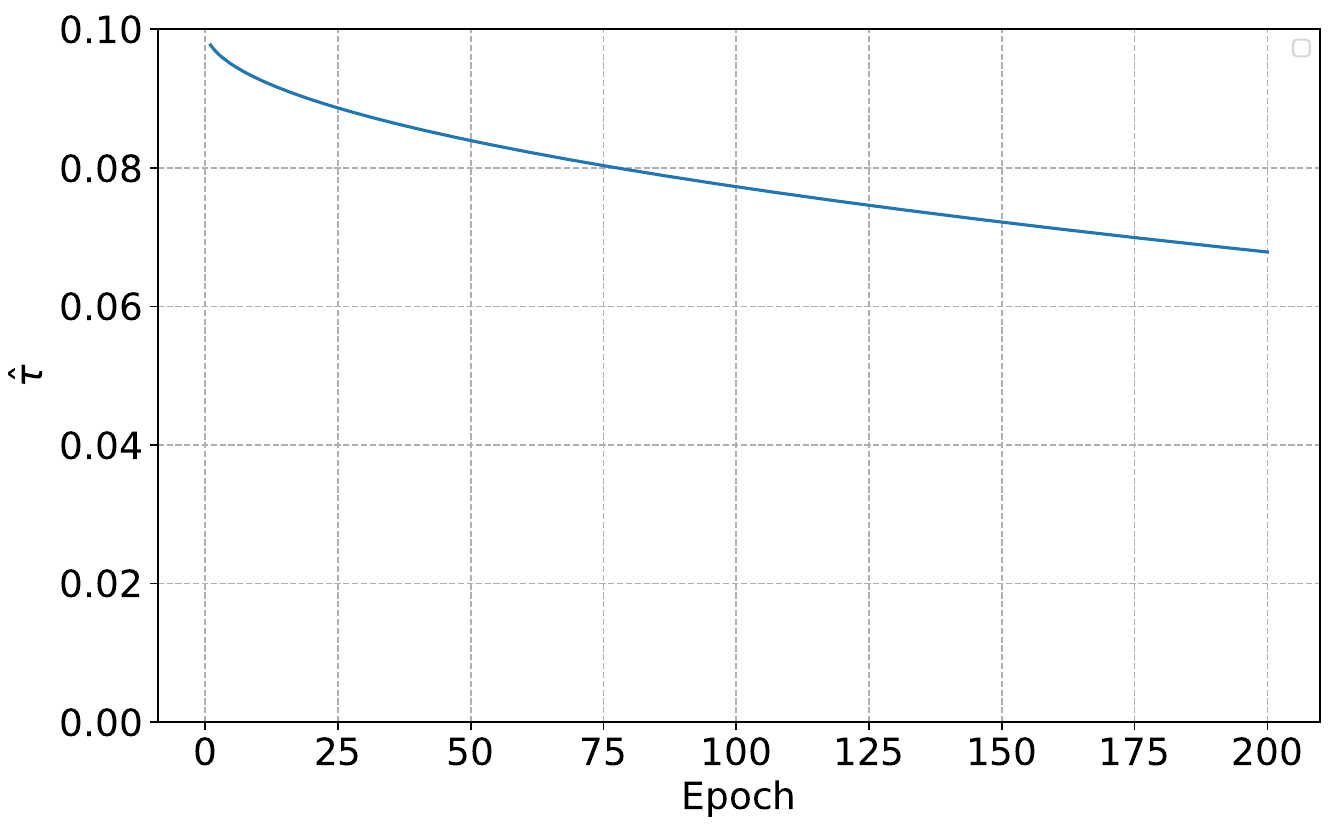}}
      \caption{Variation of temperature $\Tilde{\tau}$ for head class $n_1$ and tail class $n_5$ on CIFAR10-LT. (a) corresponds to the head class $n_1$; (b) shows the tail class $n_5$.}
	\label{fig4}

\end{figure}

\subsection{The variation of the temperature for head and tail class}
We examine the variation of the temperature on the CIFAR10-LT benchmark, where the percentage ($k$) of tail classes is set to 0.6
The head classes encompass $n_1,n_2,n_3,n_4$, while the tail classes consist of $n_5,n_6,n_7,n_8,n_9,n_{10}$. 
To comprehensively understand the temperature adjustment strategy during training, we plot the variations of the temperature value $\Tilde{\tau}$, defined in Eq.(\ref{eq4}), for the head class $n_1$ and tail class $n_5$. The default values of $\tau=0.1$ and $E=200$ are utilized for this analysis.
As depicted in Figure \ref{fig4a}, concerning the head class $n_1$, the temperature value $\Tilde{\tau}$ gradually diminishes from 0.1 to 0.01 during training. The reduction in $\Tilde{\tau}$ can enhance the pushing effect between OOD samples and head classes, thereby amplifying the disparity between head class and OOD samples. 

Similarly, as illustrated in Figure \ref{fig4b}, the temperature value $\Tilde{\tau}$ for tail class $n_5$ progressively decreases.
Smaller $\Tilde{\tau}$ reinforce the pulling effect within each tail class, thereby rendering tail classes and OOD samples more discernible. Noteworthy is the less pronounced decrease in the magnitude of $\Tilde{\tau}$ for tail class $n_5$ compared to head class $n_1$. Given the scarcity of tail class samples in the training data, an excessively low $\Tilde{\tau}$ may not effectively capture the nuanced differences between tail classes, potentially leading to a negative impact on classification accuracy.

\section{The Informative Outlier Mining Algorithm}
\label{The Informative Outlier Mining Algorithm}

Algorithm~\ref{algorithm1} outlines the informative outlier mining process. In each training epoch, for a batch of ID samples of size $\mathcal{B}$, we randomly sample $3 \times \mathcal{B}$ outliers from the auxiliary OOD training set. Each outlier is then assigned a score using the outlier scoring function defined in Eq.(\ref{eq5}), which reflects its tendency to resemble head, tail, or neither class types. We then sort these outliers in descending order by score to group them into three categories:

\begin{itemize}
    \item $\emph{Tail-class-like OOD}$: This category refers to outliers that exhibit characteristics resembling tail-class samples. If an outlier belongs to this category, it indicates a higher probability of being incorrectly predicted as tail class.
    \item $\emph{Head-class-like OOD}$: This category pertains to outliers that possess attributes similar to head-class samples. If an outlier falls into this category, it suggests an ease of being predicted as head class.
    \item  $\emph{Neutral OOD}$: Outliers in this category lack distinctive features aligning with either head or tail classes, resulting in no clear prediction bias. Their ambiguous characteristics make them particularly challenging for the model to distinguish.
\end{itemize}

\begin{algorithm}[h]
\caption{Informative Outlier Mining}
\label{algorithm1}
\begin{algorithmic}[1]
\REQUIRE {Auxiliary dataset $\mathcal{D}_{\text {out}}^{\text {train}}$, batch size $\mathcal{B}$}
\STATE Randomly sample $3 \times \mathcal{B}$ outliers from $\mathcal{D}_{\text {out}}^{\text {train}}$ to get a candidate subset $V=\left\{(\tilde{\boldsymbol{x}}_i, \tilde{y})\right\}_{i=1}^{3\mathcal{B}}$ ;
\STATE Calculate score $s_i$ for each outlier $\tilde{\boldsymbol{x}_i} \in V$ using outlier score function (in Eq.(\ref{eq5}));
\STATE Store score $s_i$ in Score Set $S=\{s_{\tilde{\boldsymbol{x}}_i}\}_{i=1}^{3\mathcal{B}}$;
\STATE Sort $S$ in descending order
\STATE  Classify top $\frac{1}{3}$ outliers into $V_{\text{tail}}$ ($\emph{Tail-class-like OOD}$);
\STATE  Classify middle $\frac{1}{3}$ outliers into $V_{\text{neutral}}$ ($\emph{Neutral OOD}$);
\STATE  Classify bottom $\frac{1}{3}$ outliers into $V_{\text{tail}}$ ($\emph{Head-class-like OOD}$);
\end{algorithmic}
\end{algorithm}

By employing informative outlier mining, we aim to identify outliers that provide valuable information for long-tailed OOD detection. This approach enables the selection of outliers that possess specific characteristics related to either tail or head classes, facilitating the improvement of OOD detection performance under long-tailed settings.

\section{The RSCL Algorithm}
\label{The RSCL Algorithm}

The complete RSCL training process is shown in Algorithm \ref{algorithm2}. 


\begin{algorithm}[h]
\caption{RSCL}
\label{algorithm2}
\begin{algorithmic}[1]
\REQUIRE {Training dataset $\mathcal{D}_{\text {in}}^{\text {train}}$, Auxiliary dataset $\mathcal{D}_{\text {out}}^{\text {train}}$, batch size $\mathcal{B}$, training epochs $E$, loss weight $\alpha$, $\beta$ and $\gamma$ }
\FOR{$e$= $1,2,...,E$}

\FOR{each iteration}
\STATE Sample $\mathcal{B}$ in-distribution data from $\mathcal{D}_{\text {in}}^{\text {train}}$;
\STATE Sample 3*$\mathcal{B}$ candidate outliers from $\mathcal{D}_{\text {out}}^{\text {train}}$, and perform informative outlier mining (refer to Algorithm \ref{algorithm1});
\IF{$e \leq \frac{3}{4}E$} 
\STATE Sample $\mathcal{B}/3$ outliers from $\emph{Tail-class-like OOD}$, $\emph{Head-class-like OOD}$, and $\emph{Neutral OOD}$ categories, respectively, to construct a subset containing $\mathcal{B}$ mixed outliers;
\STATE Perform common gradient descent on model $\boldsymbol{f}$ with $\mathcal{L}_{\text {RSCL }}$ defined in Eq.(\ref{eq6}).
\ELSE
\STATE Sample $\mathcal{B}$ outliers from the $\emph{Neutral OOD}$ category as the auxiliary OOD training data,
\STATE Perform common gradient descent on model $\boldsymbol{f}$ with $\mathcal{L}_{\text {RSCL }}$ defined in Eq.(\ref{eq6}).
\ENDIF 
\ENDFOR
\ENDFOR
\end{algorithmic}
\end{algorithm}

\section{More Experiment Settings}
\subsection{Datasets}
\label{appendix_datasets}
\textbf{In-distribution training and test sets} ($\mathcal{D}_{\text {in}}^{\text {train}}$, $\mathcal{D}_{\text {in}}^{\text {test}}$) 
\quad We use three popular long-tailed image classification datasets as $\mathcal{D}_{\text {in}}^{\text {train}}$, including CIFAR10-LT, CIFAR100-LT \cite{cao2019learning}, and ImageNet-LT \cite{liu2019large}.
The original version of CIFAR10 and CIFAR100 contains 50,000 training images and 10,000 validation images of size 32×32 with 10 and 100 classes, respectively. CIFAR10-LT and CIFAR100-LT are the imbalanced version of them, which reduce the number of training examples per class and keep the validation set unchanged. The imbalance ratio $\rho$ denotes the ratio between sample sizes of the most frequent class and least frequent class.
ImageNet-LT is a large-scale dataset in long-tail recognition, which truncates the balanced version ImageNet \cite{imagenet}. ImageNet-LT has 1,000 classes, which contain 115,846 training images with the number of per-class training data ranging from 5 to 1,280, and 20,000 validation images with a balanced class size.

\textbf{OOD training set} ($\mathcal{D}_{\text {out}}^{\text {train}}$) 
\quad TinyImages80M \cite{tinyImages80M} contains 80 million images with a size of 32×32. We use a subset of random 300K images from TinyImages80M as $\mathcal{D}_{\text {out}}^{\text {train}}$ for CIFAR10-LT and CIFAR100-LT. For ImageNet-LT, we use ImageNet-Extra as $\mathcal{D}_{\text {out}}^{\text {train}}$ following \cite{pascl}. ImageNet-Extra contains 517,711 images belonging to 500 classes from ImageNet-22k \cite{imagenet}, but having not overlapping with the 1,000 in-distribution classes in ImageNet-LT.

\begin{figure}[h]
  \centering
   \includegraphics[width=0.6\linewidth]{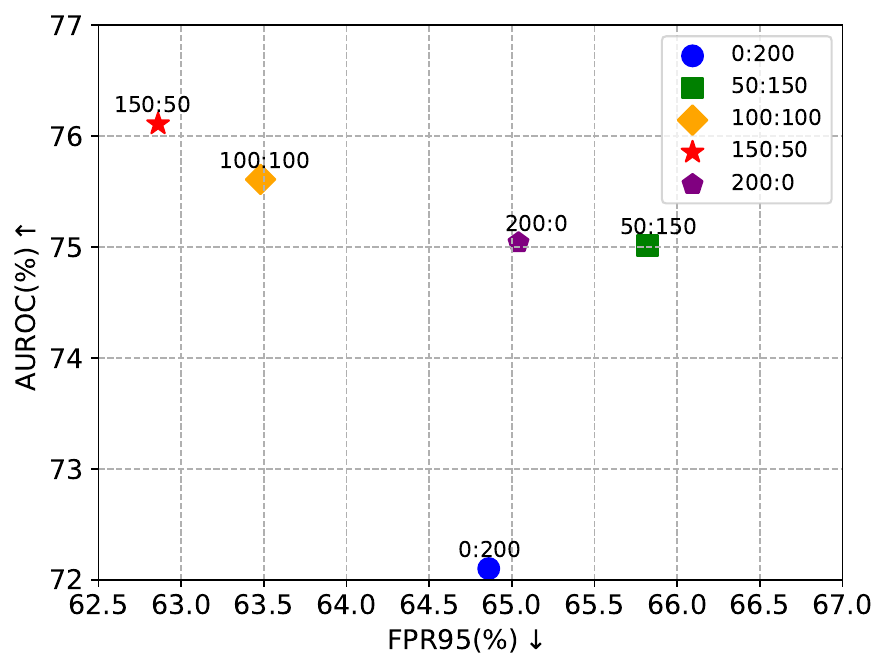}
   \caption{The evaluation of OOD detection performance through utilizing diverse types of outliers at different training epochs. The notation “150:5” denotes the method of employing a subset comprising mixed outliers from three distinct categories during the initial 150 training epochs, followed by the use of a subset exclusively consisting of neutral outliers for the subsequent 50 training epochs. The notation “0:200” signifies a method devoid of mixed outliers, relying solely on neutral outliers throughout the training process. Conversely, the notation “200:0” denotes a method focused solely on the use of mixed outliers, excluding the use of neutral outliers during training. }
   \label{fig5}
   \vspace{-5mm}
\end{figure}

\textbf{OOD test set} ($\mathcal{D}_{\text {out}}^{\text {test}}$) 
\quad We use SC-OODbenchmark \cite{yang2021semantically} as true OOD data for CIFAR10-LT and CIFAR100-LT following \cite{pascl,EAT,TSCL,COCL}. The SC-OOD benchmark contains six datasets: Texture \cite{texture}, SVHN \cite{SVHN}, CIFAR \cite{CIFAR}, Tiny ImageNet \cite{Tinyimagenet}, LSUN \cite{LSUN}, and Places365 \cite{Places365}. For ImageNet-LT, we use ImageNet-1k-OOD constructed by \cite{pascl} as $\mathcal{D}_{\text {out}}^{\text {test}}$. ImageNet-1k-OOD has 50,000 OOD test images from 1,000 classes randomly selected from ImageNet-22k \cite{imagenet} (with 50 images in each class), which is of the same size as the in-distribution test set. The 1,000 classes in ImageNet-1k-OOD are not ovelapped with either the 1,000 in-distribution classes in ImageNet-LT or the 500 OOD training classes in ImageNet-Extra.

\subsection{Implementation Details} 
\label{Implementation Details}
In this section, we provide more implementation details about our approach RSCL. For experiments on CIFAR10-LT and CIFAR100-LT, we train the ResNet18 model for 200 epochs using Adam optimizer with initial learning rate $1 \times 10^{-3}$ and batch size 256. The learning rate is decayed to 0 using a cosine annealing learning rate scheduler.
For experiments on ImageNet-LT, we follow the settings in \cite{pascl,COCL,EAT} and use ResNet50 as backbone. We train the model for 100 epochs using SGD optimizer with initial learning rate 0.1 and batch size 60. We decay the learning rate by a factor of 10 at epoch 60 and 80.
On all datasets, we set $\tau$ = 0.1 following \cite{pascl,TSCL}. Empirical analysis leads us to configure the loss weight hyperparameters as follows: the percentage of tail classes $k$ = 0.6, loss weight parameter $\alpha$ = 0.05, $\beta$ = 0.05, $\gamma$ = 0.1. 


For other hyper-parameters in the compared methods, we use the values suggested in the original papers. 
Significantly, the implementation of the original PASCL \cite{pascl} and EAT \cite{EAT} approaches incorporated fine-tuning techniques. However, as our RSCL approach does not involve fine-tuning technique, to ensure a fair comparison, we re-run the PASCL and EAT without using the fine-tuning techniques. 
Furthermore, the original COCL approach \cite{COCL} integrated a calibration technique at inference time, and the FPR95 evaluation metric used by COCL is different from ours. Therefore, we re-run the COCL approach utilizing our evaluation metrics and refrain from using the calibration technique at inference time. In particular, even when compared with the original COCL and EAT method, RSCL still achieves the best OOD detection performance. All experiments in this paper are conducted on a single NVIDIA A100 GPU.

\begin{table}[h]\centering
\caption{Ablation on the percentage ($k$) of tail classes. Experiments are conducted on CIFAR100-LT using ResNet-18. The first row ($k = 100\%$) means that $\mathcal{L}_{\text{head}}$ is not used and $\mathcal{L}_{\text{tail}}$ is applied to all ID training samples. The last row ($k = 0\%$) denotes the exclusion of $\mathcal{L}_{\text{tail}}$, with $\mathcal{L}_{\text{head}}$ being applied to all ID training samples. All values are percentages averaged over six OOD test datasets. Bold numbers are the best results.}
\label{table7}
\begin{tabular}{cccccc}
\hline  & $k$ & AUROC $\uparrow$ & AUPR $\uparrow$  & FPR95 $\downarrow$ & ACC $\uparrow$ \\
\hline  & $100 \%$ & 74.87 & 60.09 & 63.63 & 40.25  \\
\cline { 1 - 6 }  & $70 \%$ & 75.33 & 69.62 & 64.16 & \textbf{43.23}\\
& $\textbf{60\%}$ & $\textbf{76.11}$ & $\textbf{70.76}$ & $\textbf{62.86}$ & 42.81 \\
& $50 \%$ & 75.64 & 70.48 & 63.82 & 42.10 \\
& $40 \%$ & 75.25 & 70.39 & 64.75 & 41.08 \\
\hline & $0 \%$ & 75.06 & 69.72 & 64.57 & 43.02 \\

\hline 
\end{tabular}

\end{table}

\begin{table*}[]\centering
\footnotesize 
\caption{Ablation on different model structures trained on CIFAR100-LT. All values are percentages averaged over six OOD test datasets. Bold numbers are the best results.} 
\label{table8}
\begin{tabular}{cccccc}
\hline
Model     & Method      & \textbf{AUROC$\uparrow$} & \textbf{AUPR$\uparrow$} & \textbf{FPR95$\downarrow$} & \textbf{ACC$\uparrow$} \\ \hline
\multirow{6}{*}{\begin{tabular}[c]{@{}c@{}}ResNet-34
\end{tabular}} 
        
        & OE \cite{OE}          & 73.11           & 67.55          & 68.15          & 39.00   \\
        & PASCL \cite{pascl}      & 73.66          & 67.99           & 67.15          & 39.54   \\
        
        & EAT \cite{EAT} & 73.81  & 68.94  & 67.22 & \textbf{43.26} \\
        & COCL \cite{COCL} & 74.40  & \textbf{70.57}  & 68.17 & 42.06 \\
        & TSCL \cite{TSCL} & 74.57  & 68.15  & 64.59 & 40.18 \\ 
        & \textbf{RSCL(Ours)} & \textbf{75.05}  & 69.61  & \textbf{64.10} & 42.54 \\ \hline
\multirow{6}{*}{\begin{tabular}[c]{@{}c@{}}ResNet-50
\end{tabular}} 
        & OE \cite{OE}         & 72.70           & 67.62           & 69.77          & 37.86   \\
        & PASCL \cite{pascl}     & 73.07           &  68.10          &  69.63         & 37.74   \\
        
        & EAT \cite{EAT} & 73.04  & 68.56  & 68.98 & 41.21 \\
        & COCL \cite{COCL}  & 74.55  & 70.76  & 69.73 & 40.70 \\
        & TSCL \cite{TSCL}& 73.73  & 68.14  & 66.27 & 38.05 \\
        & \textbf{RSCL(Ours)} & \textbf{75.59}  & \textbf{70.96}  & \textbf{64.11} & \textbf{42.02} \\ 
        \hline
\end{tabular}
\end {table*}

\section{More Experimental Results}


\subsection{More Ablation Study}
\label{More Ablation Study}
\textbf{The advantage of utilizing diverse types of outliers at different training epochs.}
\quad As introduced in Algorithm \ref{algorithm2}, various types of outliers are utilized at different training epochs.
Figure \ref{fig5} shows a comparative evaluation of OOD detection performance concerning the combination of mixed outliers from three distinct categories and neutral outliers.
It is observed that initiating the model training with mixed outliers, followed by the utilization of neutral outliers, exhibits commendable OOD detection performance.
Furthermore, extending the training epochs with mixed outliers rather than neutral outliers yields a more favorable impact on enhancing OOD detection performance.
This mechanism leverages the model's exposure to diverse OOD characteristics through prolonged training with mixed outliers, consequently enhancing the model's fundamental discriminatory capacity.
Subsequently, the inclusion of more intricate neutral outliers steers the model towards mastering complex OOD instances, thereby enhancing the model's acuity in OOD detection.


\textbf{Abaltion on the percentage ($k$) of tail classes.}
\label{sup9.2}
\quad The percentage ($k$) of tail classes emerges as a crucial hyperparameter governing the delineating between head and tail classes. Ablation analyses concerning $k$ are consolidated in Table \ref{table7}. Optimal results materialize when $k$ is calibrated to approximately 60$\%$, a setting that aligns with the default value employed in our experimental setup. Notably, the results observed at $k = 60\%$ markedly outperform those at $k = 100\%$ (without $\mathcal{L}_{\text{head}}$) and $k = 0\%$ (without $\mathcal{L}_{\text{tail}}$), emphasizing the importance of our proposed approach in enhancing model performance. Furthermore, our approach exhibits notable robustness across a broad spectrum of $k$ values (e,g, $k \in[50 \%, 70 \%]$ ).

\textbf{Robustness under different model structures.}
\label{sup9.4}
\quad To evaluate the effectiveness of our approach across different model structures, we conduct ablation experiments on ResNet-34 and ResNet-50, with the results presented in Table \ref{table8}. Noteworthy is the consistent outperformance of our approach over OE \cite{OE}, PASCL \cite{pascl}, EAT \cite{EAT}, COCL \cite{COCL}, and TSCL \cite{TSCL} when applied to ResNet-50, as evidenced by superior OOD detection performance and ID classification accuracy. For instance, on ResNet-50, our approach achieves an improvement in average AUROC by \textbf{1.04$\%$}, a reduction in average FPR95 by \textbf{5.62$\%$}, an enhancement in ID classification accuracy by \textbf{1.32$\%$} compared to COCL. Similarly, promising trends are observed on ResNet-34. These findings underscore the robust performance of our approach, independent of the specific model structures used.

\begin{table*}[]\centering
\footnotesize
\caption{Average FPR95 on CIFAR100-LT using ResNet18 depending on hyperparmeter $\alpha$, $\beta$ and $\gamma$. All values are percentages averaged over six OOD test datasets. Bold numbers are the best results.}
\label{table9}
\begin{tabular}{cllllll}
\hline
\multicolumn{7}{c}{Average FPR95}                                                                                                                                                                         \\ \hline
\multicolumn{1}{c|}{\multirow{4}{*}{$\alpha$}} & \multicolumn{6}{c}{$\beta$}                                                                                                                                         \\ \cline{2-7} 
\multicolumn{1}{c|}{}                    & \multicolumn{2}{c|}{0.01}                               & \multicolumn{2}{c|}{0.05}                                        & \multicolumn{2}{c}{0.1}           \\ \cline{2-7} 
\multicolumn{1}{c|}{}                    & \multicolumn{6}{c}{$\gamma$}                                                                                                                                         \\ \cline{2-7} 
\multicolumn{1}{c|}{}                    & \multicolumn{1}{l|}{0.05}  & \multicolumn{1}{l|}{0.1}   & \multicolumn{1}{l|}{0.05}  & \multicolumn{1}{l|}{0.1}            & \multicolumn{1}{l|}{0.05}  & 0.1   \\ \hline
\multicolumn{1}{l|}{0.01}                & \multicolumn{1}{l|}{66.05} & \multicolumn{1}{l|}{65.48} & \multicolumn{1}{l|}{65.45} & \multicolumn{1}{l|}{65.38}          & \multicolumn{1}{l|}{65.61} & 64.69 \\ \hline
\multicolumn{1}{l|}{0.05}                & \multicolumn{1}{l|}{64.45} & \multicolumn{1}{l|}{64.94} & \multicolumn{1}{l|}{64.21} & \multicolumn{1}{l|}{\textbf{62.86}} & \multicolumn{1}{l|}{64.65} & 63.48 \\ \hline
\multicolumn{1}{l|}{0.1}                 & \multicolumn{1}{l|}{63.69} & \multicolumn{1}{l|}{63.61} & \multicolumn{1}{l|}{63.91} & \multicolumn{1}{l|}{63.69}          & \multicolumn{1}{l|}{63.43} & 63.47 \\ \hline
\end{tabular}
\end{table*}

\begin{table*}[ht]\centering
\footnotesize
\caption{Accuracy on CIFAR100-LT using ResNet18 depending on hyperparmeter $\alpha$, $\beta$ and $\gamma$. All values are percentages averaged over six OOD test datasets. Bold numbers are the best results.}
\label{table10}
\begin{tabular}{cllllll}
\hline
\multicolumn{7}{c}{ACC}                                                                                                                                                                         \\ \hline
\multicolumn{1}{c|}{\multirow{4}{*}{$\alpha$}} & \multicolumn{6}{c}{$\beta$}                                                                                                                                         \\ \cline{2-7} 
\multicolumn{1}{c|}{}                    & \multicolumn{2}{c|}{0.01}                               & \multicolumn{2}{c|}{0.05}                                        & \multicolumn{2}{c}{0.1}           \\ \cline{2-7} 
\multicolumn{1}{c|}{}                    & \multicolumn{6}{c}{$\gamma$}                                                                                                                                         \\ \cline{2-7} 
\multicolumn{1}{c|}{}                    & \multicolumn{1}{l|}{0.05}  & \multicolumn{1}{l|}{0.1}   & \multicolumn{1}{l|}{0.05}  & \multicolumn{1}{l|}{0.1}            & \multicolumn{1}{l|}{0.05}  & 0.1   \\ \hline
\multicolumn{1}{l|}{0.01}                & \multicolumn{1}{l|}{42.48} & \multicolumn{1}{l|}{\textbf{43.60}} & \multicolumn{1}{l|}{42.34} & \multicolumn{1}{l|}{43.26}          & \multicolumn{1}{l|}{41.72} & 42.82 \\ \hline
\multicolumn{1}{l|}{0.05}                & \multicolumn{1}{l|}{41.63} & \multicolumn{1}{l|}{42.83} & \multicolumn{1}{l|}{41.72} & \multicolumn{1}{l|}{42.81} & \multicolumn{1}{l|}{41.34} & 42.71 \\ \hline
\multicolumn{1}{l|}{0.1}                 & \multicolumn{1}{l|}{41.41} & \multicolumn{1}{l|}{42.33} & \multicolumn{1}{l|}{41.01} & \multicolumn{1}{l|}{42.53}          & \multicolumn{1}{l|}{40.78} & 42.30 \\ \hline
\end{tabular}
\end{table*}

\textbf{Ablation on loss weight $\alpha$, $\beta$ and $\gamma$}
\label{supc2}
\quad Table \ref{table9} and Table \ref{table10} show the impact of different loss weights on both OOD detection performance and ID classification accuracy. The hyperparameter $\alpha$ is set to strike a balance between learning from ID classes and outlier classes. A higher value assigned to $\alpha$ indicates a model preference towards outliers. Elevating $\alpha$ leads to enhanced OOD detection performance but may compromise ID classification accuracy. Hence, selecting an appropriate $\alpha$ becomes crucial. Furthermore, $\beta$ and $\gamma$ are designated to balance the  differentiation between OOD samples and head/tail samples, with a slightly higher value for $\gamma$ compared to $\beta$ often proving advantageous for both OOD detection and ID classification. Finally, a configuration where $\alpha$ = 0.05, accompanied by corresponding values of $\beta$ = 0.05 and $\gamma$ = 0.1, is generally recommended and serves as the default setting in our experiment settings.

\textbf{Robustness under different imbalance ratios ($\rho$).} 
\quad We have showcased the excellent performance of RSCL on $\rho$ = 100 in Section \ref{Main Results and Discussion}. Furthermore, RSCL exhibits robust performance across various imbalance ratios such as $\rho$ = 50 and $\rho$ = 20. The results in Table \ref{table11} underscore that RSCL consistently outperforms OE and PASCL by a considerable margin in terms of OOD detection performance and ID classification accuracy. Moreover, RSCL demonstrates superior performance compared to recent methods such as TSCL, EAT, and COCL.

\begin{table}[h]
\centering
\small 
\caption{Robustness under different imbalance ratios ($\rho$). All values are percentages averaged over six OOD test datasets. Bold numbers are the best results. } 
\label{table11}
\begin{tabular}{cccccc}
\hline
Imbalance Ratio     & Method      & \textbf{AUROC$\uparrow$} & \textbf{AUPR$\uparrow$} & \textbf{FPR95$\downarrow$} & \textbf{ACC$\uparrow$} \\ \hline
\multirow{6}{*}{\begin{tabular}[c]{@{}c@{}}$\rho=50$
\end{tabular}} 
        
        & OE          & 74.07           & 68.38          & 66.68          & 43.47   \\
        & PASCL      & 74.35           & 68.51           & 66.49          & 44.00   \\
        & EAT & 75.45  & 70.02  & 63.62 & 46.17 \\
        & COCL & 76.07  & 71.42  & 65.49 & 46.61 \\
        & TSCL & 75.59  & 69.34  & 62.95 & 44.22 \\ 
        & \textbf{RSCL (Ours)} & \textbf{76.50}  & \textbf{70.79}  & \textbf{61.56} & \textbf{46.64} \\ \hline
\multirow{6}{*}{\begin{tabular}[c]{@{}c@{}}$\rho=20$
\end{tabular}} 
        & OE          & 76.51           & 70.57           & 63.06          & 50.98   \\
        & PASCL      & 77.34           &  70.80          &  61.81         & 50.13   \\
        & EAT & 77.43  & 72.47  & 62.14 & 53.76 \\
        & COCL & 78.52  & \textbf{73.27}  & 59.94 & 53.67 \\
        & TSCL & 77.64  & 71.46  & 60.11 & 50.69 \\
        & \textbf{RSCL (Ours)} & \textbf{78.71}  & 73.19  & \textbf{58.12} & \textbf{53.80} \\ 
        \hline
\end{tabular}

\end {table}

\section{Limitation and Broader Impacts}
\label{Limitation and Broader Impacts}
\subsection{Limitation}
\label{Limitation}
While RSCL offers a conceptually simple and empirically effective framework for out-of-distribution detection under long-tailed distributions, it is primarily designed for specific types of classification scenarios and may not directly generalize to all OOD detection settings.

First, RSCL assumes a fixed set of imbalanced in-distribution (ID) classes, where class-wise temperature adjustment and informative outlier mining can be effectively leveraged. Its performance gains are most pronounced in such long-tailed classification tasks. In more balanced datasets or tasks with fewer classes, these mechanisms may provide limited additional benefit.

Second, the current framework relies on access to auxiliary OOD data to mine semantically informative outliers. While such data are readily available in many benchmark and real-world settings, RSCL does not directly address scenarios where external OOD sources are scarce or unavailable. In such cases, future extensions could integrate synthetic outlier generation, unsupervised pretraining, or generative modeling to mitigate this dependency.

Finally, RSCL focuses on single-stage offline training under a static label space. More dynamic or open-world settings, such as continual learning or open-vocabulary OOD detection, remain beyond the current scope and provide promising directions for future research.

\subsection{Broader Impacts}
\label{Broader Impacts}

Long-tailed OOD detection aims to improve the reliability of deep learning models in imbalanced real-world environments, where rare but critical classes are often underrepresented. Our proposed method, RSCL, is designed to enhance OOD detection in such settings by improving the model's ability to distinguish among head classes, tail classes, and out-of-distribution samples. This contributes positively to the safety and robustness of AI systems in high-stakes applications such as medical diagnosis, fraud detection, and autonomous systems, where both rare event recognition and anomaly detection are crucial. When deploying RSCL, it is important to ensure that the method is used solely for improving model reliability and fairness in safety-critical domains, and not in ways that compromise individual privacy, reinforce harmful biases, or support surveillance without consent.


\clearpage

\end{document}